\pdfoutput=1
\documentclass{article}

\usepackage{fullpage}
\usepackage{authblk}
\usepackage{natbib}
\usepackage[utf8]{inputenc} 
\usepackage[T1]{fontenc}
\usepackage{url}
\usepackage{hyperref}       
\usepackage{booktabs}       
\usepackage{amsfonts}       
\usepackage{nicefrac}       
\usepackage{microtype}      
\usepackage{ifthen}
\usepackage[linesnumbered,ruled]{algorithm2e}
\usepackage[cmex10]{amsmath}
\usepackage{amssymb, bbm, bm}
\usepackage{graphicx}
\usepackage{subfig}
\usepackage{wrapfig}
\usepackage{cite}
\usepackage[font=small]{caption}
\usepackage[colorinlistoftodos,prependcaption]{todonotes}

\usepackage{fancyref}

\makeatletter
\let\saved@reftextfaraway\reftextfaraway
\renewcommand*{\reftextfaraway}[1]{%
  \begingroup
    \def\ref@unknown@value{??}%
    \ifx\@tempa\ref@unknown@value
      \count@=0 %
    \else
      \count@\thevpagerefnum\relax
      \advance\count@ by -\@tempa\relax
      \ifnum\count@<0 \count@=-\count@\fi
    \fi
    \ifnum\count@<10 %
      \unskip
    \else
      \saved@reftextfaraway{#1}%
    \fi
  \endgroup
}
\makeatother

\newtheorem{prop}{Proposition}

\newcommand{\norm}[1]{\left\lVert#1\right\rVert^2}

\newcommand{\alldata}{\mathbf{X}}
\newcommand{\onedata}{\mathbf{x}}
\newcommand{\dataspace}{\mathcal{X}}
\newcommand{\alltarget}{\mathbf{y}}
\newcommand{\onetarget}{y}

\newcommand{\randlatent}{\mathbf{Z}}
\newcommand{\onelatent}{\mathbf{z}}
\newcommand{\latentspace}{\mathcal{Z}}
\newcommand{\noise}{\eta}
\newcommand{\mean}{\mu}
\newcommand{\meanvect}{\bm{\mu}}
\newcommand{\stdev}{\sigma}

\newcommand{\kernel}{K}
\newcommand{\kernmat}{\mathbf{K}}
\newcommand{\smallkernel}{k}
\newcommand{\wtkernel}{\kappa}
\newcommand{\rffwt}{\mathbf{v}}
\newcommand{\rffwtmat}{\mathbf{V}}

\newcommand{\rfffunc}{\mathbf{f}}
\newcommand{\latfunc}{g}

\newcommand{\randgpfunc}{H}
\newcommand{\probfunc}{p}
\newcommand{\probfuncspace}{\mathcal{P}}
\newcommand{\rkhsvect}{\bm{\omega}}
\newcommand{\numtrain}{n}
\newcommand{\numlab}{n_l}
\newcommand{\numunlab}{n_u}
\newcommand{\highdim}{D}
\newcommand{\lowdim}{d}

\newcommand{\negll}{L}

\newcommand{\weights}{\mathbf{w}}
\newcommand{\randweights}{\mathbf{W}}
\newcommand{\tfweights}{\mathbf{u}}
\newcommand{\randtfweights}{\mathbf{U}}
\newcommand{\numweights}{m}
\newcommand{\numfeat}{R}
\newcommand{\featind}{r}
\newcommand{\classwtprior}{q}
\newcommand{\numclasses}{C}
\newcommand{\rkhs}{\mathcal{H}}
\newcommand{\tf}{T}
\newcommand{\shift}{s}
\newcommand{\classprob}{\rho}

\newcommand{\classind}{c}
\newcommand{\classweight}{\theta}

\newcommand{\altshift}{r}

\newcommand{\GP}{\mathcal{GP}}

\graphicspath{{Figures/}}

\DeclareMathOperator*{\argmin}{arg\,min}

\title{Deep Kernels with Probabilistic Embeddings for Small-Data Learning}

\author[1]{Ankur Mallick\thanks{Correspondence to amallic1@andrew.cmu.edu}}
\author[2]{Chaitanya Dwivedi\thanks{Work done while at Carnegie Mellon University}}
\author[3]{Bhavya Kailkhura}
\author[1]{\\Gauri Joshi}
\author[3]{T. Yong-Jin Han}
\affil[1]{Carnegie Mellon University}
\affil[2]{PathAI}
\affil[3]{Lawrence Livermore National Laboratories}
\date{}

\begin{document}

\maketitle

\begin{abstract}
Gaussian Processes (GPs) are known to provide accurate predictions and uncertainty estimates even with small amounts of labeled data by capturing similarity between data points through their kernel function. However traditional GP kernels are not very effective at capturing similarity between high dimensional data points. Neural networks can be used to learn good representations that encode intricate structures in high dimensional data, and can be used as inputs to the GP kernel. However the huge data requirement of neural networks makes this approach ineffective in small data settings. To solves the conflicting problems of representation learning and data efficiency, we propose to learn deep kernels on probabilistic embeddings by using a probabilistic neural network. Our approach maps high-dimensional data to a probability distribution in a low dimensional subspace and then computes a kernel between these distributions to capture similarity. To enable end-to-end learning, we derive a functional gradient descent procedure for training the model. Experiments on a variety of datasets show that our approach outperforms the state-of-the-art in GP kernel learning in both supervised and semi-supervised settings. We also extend our approach to other small-data paradigms such as few-shot classification where it outperforms previous approaches on \emph{mini}-Imagenet and CUB datasets.
\end{abstract}

\section{Introduction}\label{sec:intro}

\begin{figure*}[htb]
  \centering
    \subfloat[Accurate Prediction]{\includegraphics[width=0.32\linewidth]{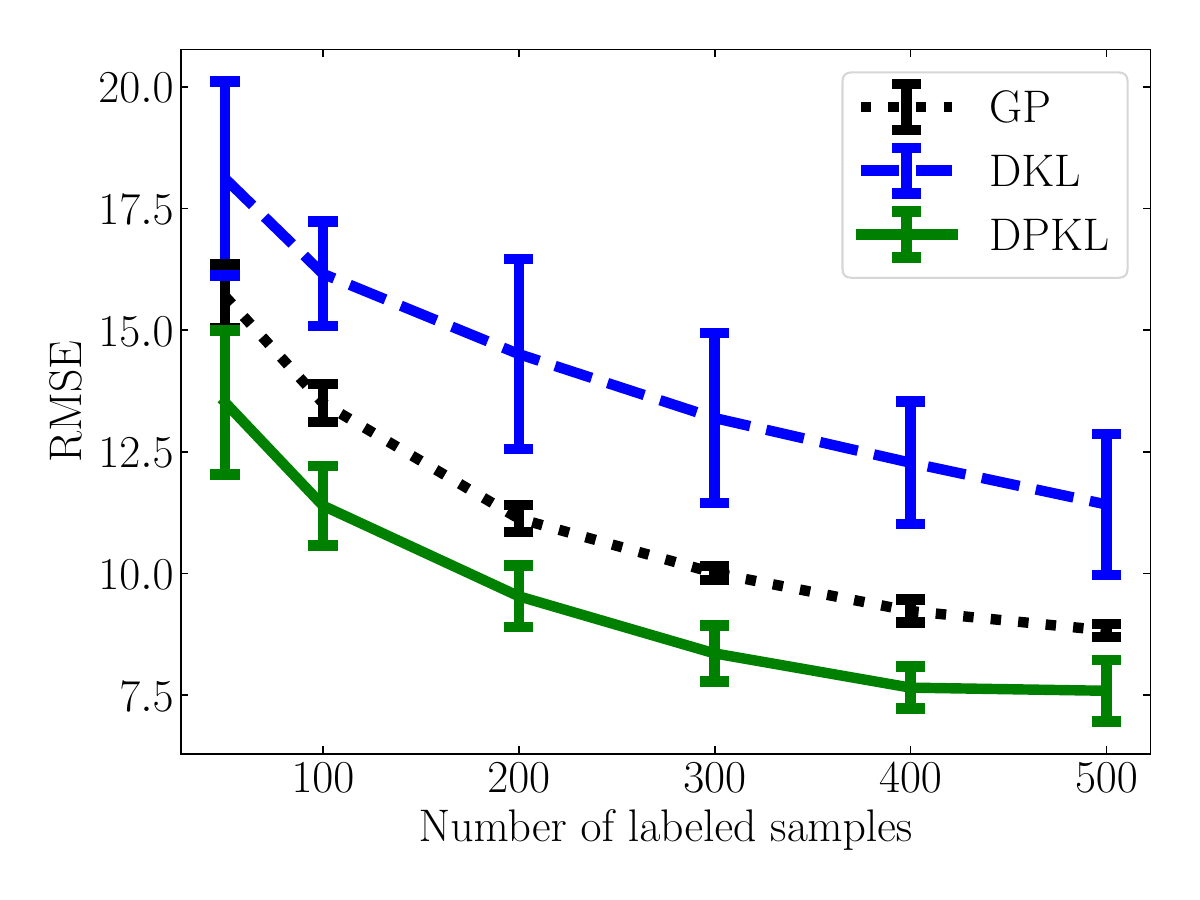}\label{fig:rmse}}
   \subfloat[Uncertainty Quantification]{\includegraphics[width=0.32\linewidth]{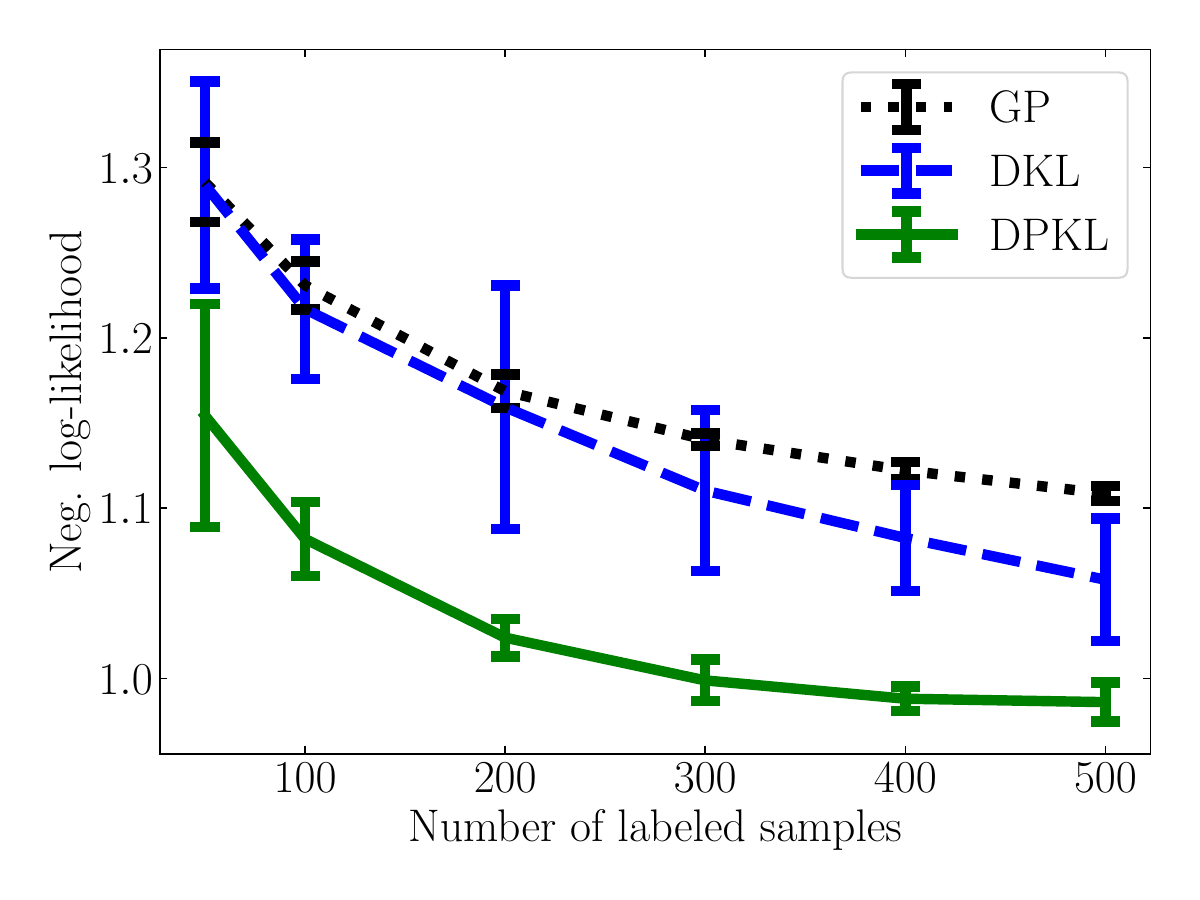}\label{fig:uq}}
   \subfloat[Representation Learning]{\includegraphics[width=0.34\linewidth]{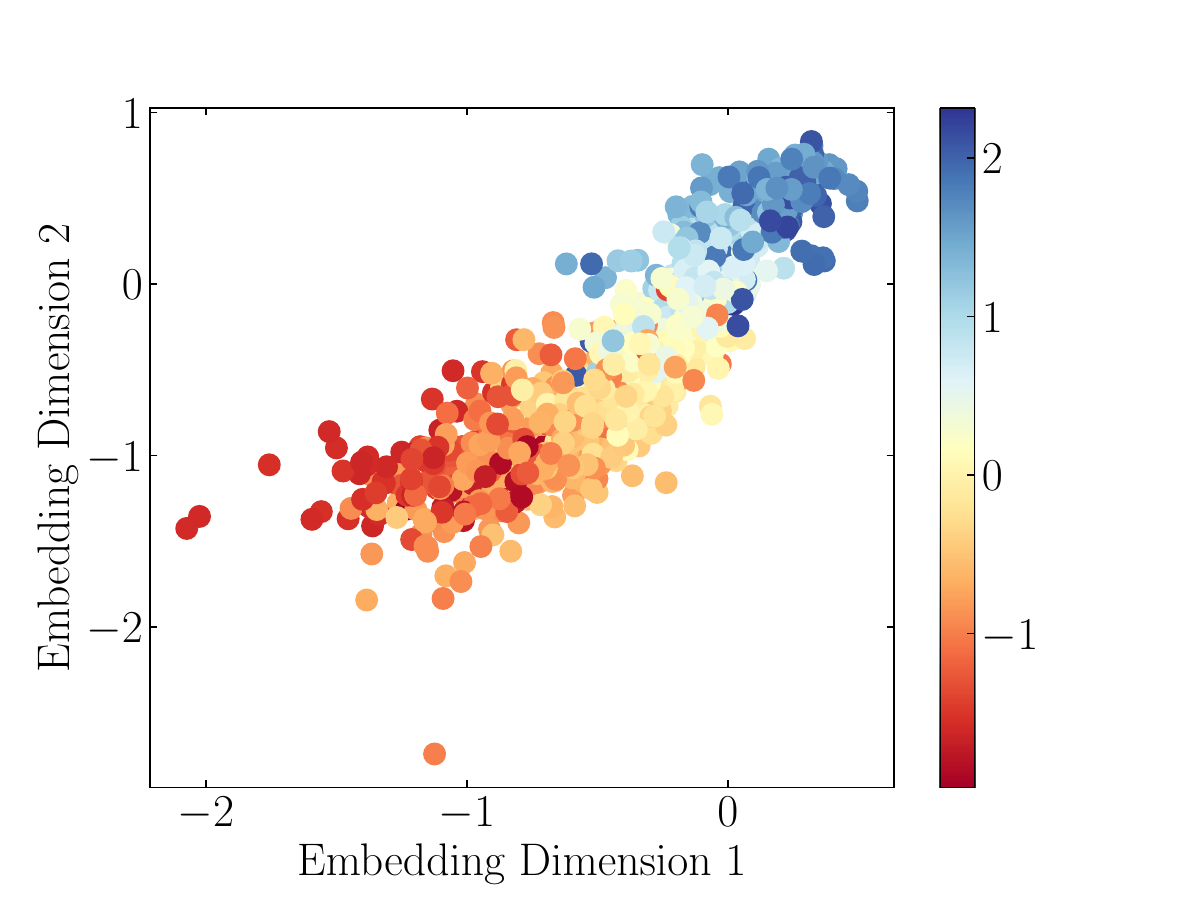}\label{fig:rl}}
   \caption{ Performance of our models for regression on the UCI CTSlice Dataset (Dimensionality = 384) \citep{graf20112d} with $\numtrain = \{50, 100, 200, 300, 400, 500\}$ labeled data points.  (a) Our approach DPKL, has \emph{significantly lower} average RMSE than 
   conventional GPs and Deep Kernel Learning (DKL). (b) DPKL not only has lower error, but also captures the data distribution much better (lower negative log likelihood) than DKL. (c) Points with similar target values have similar mean embeddings in the 2-dimensional latent space learned by DPKL with $\numtrain = 100$ labeled samples.}
  \label{fig:mainresults}
\end{figure*}

The availability of huge labeled datasets \citep{deng2009imagenet} has played a key role in the success of deep learning in recent time. Given a sufficient amount of labeled (training) data, deep neural networks achieve superhuman prediction performance on unseen (test) data \citep{krizhevsky2012imagenet,graves2013speech}. However, machine learning is increasingly being applied to areas where it is challenging to procure large amounts of labeled data, e.g., materials science \citep{zhang2020leveraging}, poverty prediction \citep{jean2016combining}, and infrastructure assessment \citep{oshri2018infrastructure}. The accuracy of most machine learning models deteriorates when trained on small datasets. An additional drawback of deep neural networks is the inability to provide accurate uncertainty estimates leading to over-confident but incorrect predictions~\citep{bulusu2020anomalous} especially when trained on small data.

Gaussian Processes (GPs) \citep{rasmussen2003gaussian} are non-parametric models that leverage correlation (similarity) between data points to give a probabilistic estimate of the target function values. Exploiting similarity in a non-parametric fashion contributes to sample efficiency, while probabilistic estimation helps to quantify model uncertainty in unlabeled regions of the input space. 
However, owing to the curse of dimensionality, the number of points required to model the covariance using traditional GP kernels grows exponentially with data dimension \citep{tripathy2016gaussian} which reduces their efficacy in the \emph{small-data} regime for high-dimensional data. The success of deep neural networks in learning low dimensional representations of high dimensional data has led to works \citep{wilson2016deep} that use a neural network to map data to a low dimensional latent space and then build a GP for prediction on the \emph{latent space}. However, owing to the huge data requirement of neural networks, these approaches are also not as effective in learning latent representations in the \emph{small-data regime}.

To enable powerful, data efficient representation learning, we propose a new approach for learning highly expressive GP kernels with small amounts of labeled data by mapping data points to \emph{probability distributions} in a latent space with a probabilistic neural network \citep{neal2012bayesian}. We then leverage the theory of kernel embeddings of distributions \citep{muandet2017kernel} to build expressive kernels, and consequently GPs, on the latent probability distributions.

Probabilistic models are known to outperform deterministic models when the amount of data is insufficient to capture the complexity of the task \citep{wilson2020bayesian}. Intuitively, we expect the uncertainty due to the model's probabilistic nature to improve generalization. This is already seen in the superiority of soft thresholding over hard thresholding in clustering, and in the superior prediction performance of Bayesian Neural Networks (BNNs) over deterministic neural networks in small data settings \citep{gal2016uncertainty}, and we expect to obtain similar benefits on using probabilistic neural networks for GP kernel learning. 

Note that our aims differ from works like \citep{graves2011practical} that train BNNs for prediction. They use the probabilistic nature of the model to capture uncertainty in prediction at the \emph{output} while we use probabilistic neural networks to improve the \emph{learned representation} which can serve as input to \emph{any predictive model}. While we mainly focus on improving \emph{GP regression} through this, we also show improvements in classification using Bayesian logistic regression \citep{jaakkola1997bayesian} on the learned probabilistic representations to illustrate the flexibility of our approach. 

The difference in aims also leads to a difference in training. While BNNs are typically trained by approximating posterior distribution over model parameters, we learn the distribution over model parameters via \emph{Maximum Likelihood Estimation} (MLE) in the space of probability distributions. MLE in our model is performed using functional gradient descent, which is not only a natural approach to perform optimization in the space of functions (probability distributions), but has also recently been shown \citep{liu2016stein} to achieve an effective middle ground between the high computational cost of MCMC \citep{andrieu2003introduction} and the poor approximation quality of Variational Inference (VI) \citep{blei2017variational} for training probabilistic models. 

\Fref{fig:mainresults} shows the results of implementing our model, Deep Probabilistic Kernel Learning (DPKL), and baseline approaches - Deep Kernel Learning (DKL) \citep{wilson2015kernel}, and a GP with a Squared Exponential (SE) kernel, on a regression task of predicting location of CT slice images on the axial axis of the human body \citep{graf20112d} given a $384-$dimensional feature vector extracted from the image. We use Root Mean Squared Error (RMSE) (\Fref{fig:rmse}) to measure prediction accuracy, and negative log-likelihood (\Fref{fig:uq}) to measure uncertainty quantification (lower is better for both). Our model \emph{outperforms both baselines} on both counts and also learns embeddings that clearly capture similarity between data points with respect to target values (\Fref{fig:rl}).
  
We also extend our model to two other paradigms in small data settings -- semi-supervised learning \citep{zhu2009introduction} and few-shot classification \citep{chen2019closer}. The former involves adding a regularizer to DPKL that provably incorporates additional structural information from unlabeled data, while the latter 
uses the learned probabilistic embedding to improve the performance of Bayesian Logistic Regression for classification especially for settings like few-shot learning where we only have a small number of training examples of each class.

\section{Background and Related Work}
\label{sec:prob}

Given $\numtrain$ training data points, $\alldata \in \dataspace \subseteq \mathbb{R}^{\numtrain \times \highdim}$, and targets, $\alltarget \in \mathbb{R}^{\numtrain \times 1}$, our goal is to accurately predict targets $\onetarget_{*}$ for unseen (test) data points $\onedata_{*}$, assuming small training set size (small $\numtrain$) and high dimensionality (large $\highdim$).

\textbf{Gaussian Processes.} Gaussian Processes (GPs) \citep{rasmussen2003gaussian} are non-parametric models that, when appropriately designed, can give high quality predictions and uncertainty estimates in the small data (small $\numtrain$) regime. A GP defines a \emph{probability distribution} over functions $\randgpfunc: \mathbb{R}^{\lowdim} \rightarrow \mathbb{R}$ such that individual function values form a multivariate Gaussian distribution i.e. $\randgpfunc(\alldata) = [\randgpfunc(\onedata_1),\ldots,\randgpfunc(\onedata_\numtrain)] \sim \mathcal{N}(\meanvect,\kernmat)
$ with entries of the mean vector given by $\mean_i = \mean(\onedata_i)$, and of the covariance matrix given by $\kernel_{ij} = \kernel(\onedata_i,\onedata_j)$. A GP is fully specified by the mean function $\mean$ (typically $\mu \equiv 0$) and the covariance kernel $\kernel$ (For eg. SE or Matern) which measures similarity between data points. Data is also assumed to be corrupted by noise $\noise \sim \mathcal{N}(0,\stdev_{\eta}^2)$ and the overall covariance is 
\begin{align*}
    \text{Cov}(H(\onedata_i),H(\onedata_j)) = \kernel(\onedata_i,\onedata_j) + \stdev_{\eta}^2\mathbbm{1}\{i=j\}
\end{align*}
Given the training data $\alldata$ and associated targets $\alltarget$, the predicted target function value $\randgpfunc(\onedata_{*})$ at an unlabeled data point $\onedata_{*}$ follows a Gaussian posterior distribution, i.e. $\Pr(\randgpfunc(\onedata_{*})|\alldata,\alltarget,\onedata_{*}) = \mathcal{N}(\mu(\onedata_{*}),\stdev^2(\onedata_{*}))$  where $\mu(\onedata_{*}) = \mathbf{\smallkernel_{*}}^{T}(\kernmat + \stdev_{\eta}^{2})^{-1}\alltarget$ and $\stdev^{2}(\onedata_{*}) = \smallkernel_{**} - \mathbf{\smallkernel_{*}}^{T}(\kernmat + \stdev_{\eta}^{2})^{-1}\mathbf{\smallkernel_{*}}$ with $\mathbf{\smallkernel_{*}} = \kernel(\alldata,\onedata_{*})$ and $\smallkernel_{**} = \kernel(\onedata_{*},\onedata_{*})$. We refer readers to \citep{rasmussen2003gaussian} for further details.

\textbf{Deep Kernel Learning.} Deep Kernel Learning (DKL) \citep{wilson2016deep,wilson2015kernel,al2016learning} uses a neural network to learn a GP kernel as $\kernmat_{ij} = \kernel(\latfunc_{\weights}(\onedata_i),\latfunc_{\weights}(\onedata_j))$ where $\kernel$ is a standard GP kernel and $\latfunc_{\weights}(.)$ represents the neural network with parameters $\weights$ which can be learned by minimizing the GP negative log likelihood, i.e.,
\begin{align}
    \weights^{*} = \argmin_{\weights} (-\log \Pr(\alltarget \mid \alldata,\weights)), \quad \alltarget \sim \mathcal{N}(\meanvect,\kernmat) \label{eq:DKL_loss}
\end{align}
Standard kernels (like SE or Matern) make simplifying assumptions on the structure and smoothness of functions which do not hold in high dimensions due to the curse of dimensionality  \citep{garnett2013active,rana2017high}. DKL avoids these assumptions by mapping high-dimensional data to low-dimensional embeddings using a neural network. But, like all neural network based models, it requires a lot of labeled data for accurate prediction.

\textbf{Probabilistic Neural Networks.} Probabilistic models learn a distribution over model parameters thus incorporating uncertainty in predictive modeling. The implicit regularization due to the probabilistic nature can improve generalization performance particularly when there is insufficient data. Prior works on Probabilistic or Bayesian Neural Networks (BNNs) \citep{neal2012bayesian,graves2011practical,hernandez2015probabilistic} seek to learn the posterior distribution over model parameters that best explains the data. While these works use the BNN to directly predict the data, we use a probabilistic neural network to improve the quality of low dimensional embeddings of high dimensional data. The probabilistic embeddings given by our model serve as inputs to predictive models like GPs. Thus our approach aims to improve GP kernel learning and not replace BNNs. Moreover BNNs do not capture \emph{similarity} between data points or provide closed form uncertainty estimates on test data like GPs do. Hence GPs and BNNs are generally not used for the same tasks. 

\textbf{Semi-Supervised Learning.} Recently, \citep{jean2018semi} proposed Semi-Supervised Deep Kernel Learning (SSDKL) approach which uses posterior regularization, given a large amount of unlabeled data, to reduce the labeled data requirement for GP kernel learning. Various other semi-supervised learning approaches \citep{kingma2014semi,odena2016semi,tarvainen2017mean,laine2016temporal} have also been studied which yield accurate predictions with few labeled and many unlabeled examples. For such settings, we propose a semi-supervised model, SSDPKL in \fref{sec:DPKL} which is constrained to ensure that the unlabeled data is not projected too far away from the labeled data in the latent space, and outperforms SSDKL for regression (\fref{sec:Expts}).

\textbf{Other related work.} There is a line of work on Deep GPs \citep{damianou2013deep,salimbeni2017doubly} seeks to improve the performance of GPs in the \emph{big-data} regime by hierarchically stacking multiple GPs and training the model using Variational Inference. The authors of \citep{salimbeni2017doubly}  observe that DeepGP is not better than a single layer GP for small datasets (with similar or more labeled examples than what we have considered) while we show that DPKL generally has lower RMSE than GP in these settings. Regression on distribution inputs has been theoretically studied in \citep{poczos2013distribution,bachoc2017gaussian,bachoc2018gaussian}. They assume that the input itself is a distribution while we use neural networks to learn distributional embeddings of deterministic inputs. There has also been a surge of interest in classification in the small-data regime or few-shot classification. Recent results are summarised in \citep{chen2019closer}. In particular, the baseline models considered in their work can be directly augmented with DPKL to improve predictive performance in the few-shot setting as we show in \Fref{sec:Expts}.

\begin{figure*}[t]
  \centering
    \subfloat[DPKL]{\includegraphics[width=0.59\linewidth]{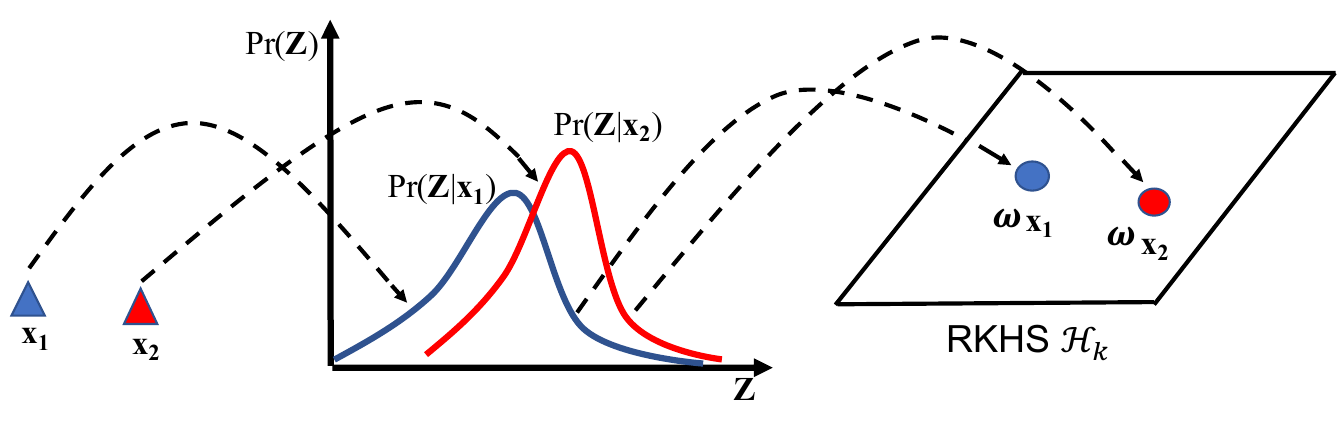}\label{fig:DPKL}}
  \subfloat[SSDPKL]{\includegraphics[width=0.39\linewidth]{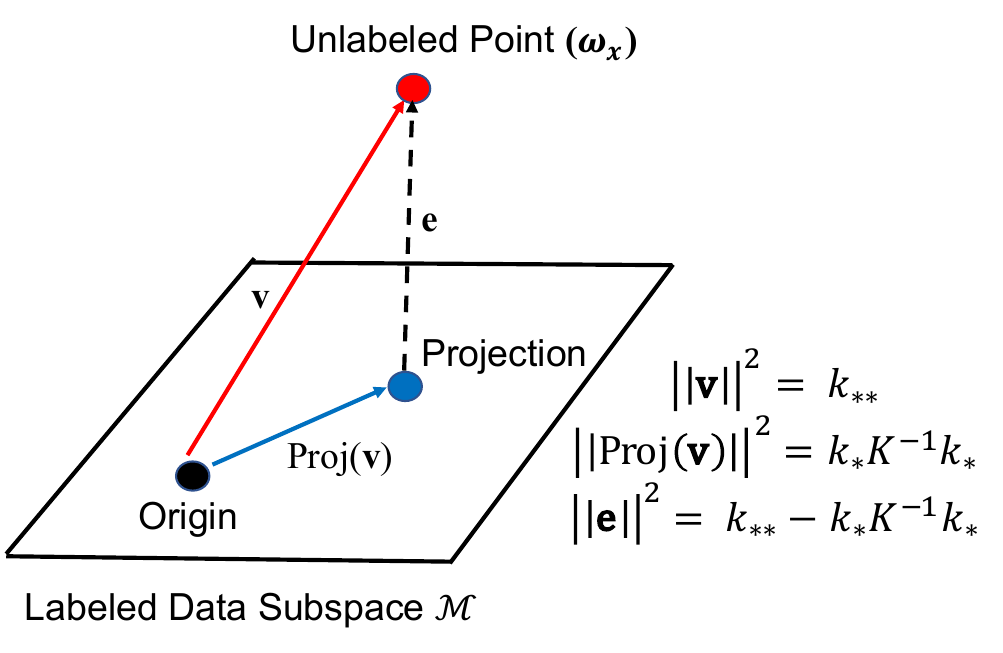} \label{fig:SSDPKL}}
   \caption{(a) Our model (DPKL) maps training points to probability distributions in the latent space and learns a GP whose kernel is the inner product between the RKHS embeddings of these distributions. (b) If unlabeled data is available we train a semi-supervised model (SSDPKL) which is a regularized version of DPKL. The regularizer is the average squared error between the RKHS norm of unlabeled points and their projections onto the subspace spanned by the labeled points.}
  \label{fig:model}
\end{figure*}

\section{Deep Probabilistic Kernel Learning}
\label{sec:DPKL}
\begin{algorithm}
\caption{Deep Probabilistic Kernel Learning \label{algo:DPKL}}
\SetKwInOut{Input}{Input}
\SetKwInOut{Output}{Output}
\Input{Training points $\alldata$ and targets $\alltarget$ along with a set of initial model parameters $\{\weights_i^{(0)}\}_{i=1}^{\numweights}$}
\Output{A set of model parameters $\{\weights_i\}_{i=1}^{\numweights} \sim \hat{\probfunc}(\weights)$ where $\hat{\probfunc}(\weights)$ is obtained by performing functional gradient descent on \fref{eq:DPKL}}
Initialize model parameters $\{\weights_i^{(0)}\}_{i=1}^{\numweights} \sim \probfunc_{0}(\weights)$ for some known $\probfunc_{0}(\weights)$\\
\For{iteration $t$}{
$\weights_{i}^{(t+1)} = \weights_{i}^{(t)} - \epsilon_{t}\phi(\weights_{i}^{(t)})$,\\
where $\phi(\weights) = \sum_{l = 1}^{\numweights}\wtkernel(\weights,\weights_l)\nabla_{\weights_l}\hat{\negll}(\weights_1,\ldots,\weights_\numweights)$
}
\end{algorithm}
In this section we first describe our model - Deep Probabilistic Kernel Learning (DPKL) illustrated in \Fref{fig:DPKL} in the context of GP regression, and then explain extensions to classification and semi-supervised learning. The proofs of theoretical results are contained in Appendix A.  

\subsection{Model Overview} 
\label{sec:overview}

DPKL predicts output $\onetarget \in \mathbb{R}$ given input $\onedata \in \dataspace \subseteq \mathbb{R}^{\highdim}$ using a Gaussian Process (GP) learned over \emph{probability distributions} in a low dimensional latent space $\latentspace \subseteq \mathbb{R}^{\lowdim}$ where distributions in $\latentspace$ are obtained by passing the input data through a probabilistic neural network.  The dimensionality $\lowdim$ of $\latentspace$ is \emph{much smaller} than the dimensionality $\highdim$ of $\dataspace$. 

\textbf{Probabilistic Latent Space Mapping.} Each data point $\onedata$ passes through a probabilistic neural network with parameters $\randweights \sim \probfunc(\randweights)$ to give a random variable $\randlatent = \latfunc_\randweights(\onedata) \in \latentspace$. Thus a point $\onedata_i$ is represented by the latent distribution $\probfunc(\randlatent|\onedata_i)$ due to the stochasticity of $\randweights$.

\textbf{GP Regression over Latent Distributions.} We assume a GP prior $\randgpfunc \sim \GP(0,\kernmat)$ over target functions $\randgpfunc$ which take \emph{distributions} $\probfunc(\randlatent|\onedata_i)$ as input and output $\hat{\onetarget_i}$, an estimate of the true label $\onetarget_i$. $\kernmat$ is the kernel matrix which models covariance between GP inputs (probability distributions). Given any two probability distributions $\probfunc(\randlatent|\onedata_i)$ and  $\probfunc(\randlatent|\onedata_j)$, the corresponding entry of $\kernmat$ is given by
\begin{align}
    \kernel_{ij} = \mathbb{E}_{\onelatent \sim \probfunc(\randlatent|\onedata_i),\ \onelatent' \sim \probfunc(\randlatent|\onedata_j)}[\smallkernel(\onelatent,\onelatent')] \label{eq:kernij}
\end{align}
Here $\smallkernel$ can be any standard kernel like the SE kernel for which the above expression would be 
\begin{align}
    \kernel_{ij} = \mathbb{E}_{\onelatent \sim \Pr(\randlatent|\onedata_i),\ \onelatent' \sim \Pr(\randlatent|\onedata_j)}[\exp(-\frac{1}{2}||\onelatent-\onelatent'||^{2})] \label{eq:kernij_rbf}
\end{align}
This choice of kernel between probability distributions is inspired by the Set Kernel of \citep{gartner2002multi} for which consistency results and minimax rates were given in \citep{szabo2016learning}. Previous works \citep{muandet2012learning,muandet2017kernel} have used similar kernels to build predictive models, such as SVMs, with distribution inputs. $\kernel_{ij}$ corresponds to the inner product between distributions $\probfunc(\randlatent|\onedata_i)$ and $\probfunc(\randlatent|\onedata_j)$ in the Reproducing Kernel Hilbert Space (RKHS) $\rkhs_\smallkernel$ defined by the kernel $\smallkernel$ \citep{muandet2017kernel} and thus $\kernmat$ is positive definite. We expect that: a) The \emph{lower dimensionality} of $\latentspace$ will improve the performance of simple kernels like SE in capturing the covariance (similarity) between points. b) \emph{Probability distributions} in the latent space $\latentspace$ will better capture uncertainty due to scarcity of training data than point embeddings. Hence we perform GP regression over distributions in $\latentspace$.

\textbf{Approximation with Random Fourier Features.} Observe that in \fref{eq:kernij} the complexity of computing $\kernel_{ij}$ is $\mathcal{O}(\numweights^2)$ (if we use $\numweights$ samples of $\onelatent$ and $\onelatent'$ to approximate the expectation). Since the total number of $(i,j)$ pairs is $\numtrain^2$ the overall complexity will be too large ($\mathcal{O}(\numtrain^2 \numweights^2)$). However, if $\smallkernel$ is shift-invariant, we can use the Random Fourier Features approximation \citep{rahimi2008random} to approximate it as
\begin{align}
    \smallkernel(\onelatent,\onelatent') \approx \rfffunc_{\rffwtmat}(\onelatent)^{T}\rfffunc_{\rffwtmat}(\onelatent')
\end{align}
where $\rffwtmat$ is a matrix whose rows $\rffwt_1,\ldots,\rffwt_\numfeat$ are i.i.d samples from the Fourier Transform of $\smallkernel$ and $\rfffunc_{\rffwtmat}(\onelatent)$ is a vector with $2\numfeat$ elements whose $l^{th}$ element is given by
\begin{align}
    \rfffunc_{\rffwtmat}(\onelatent)_\featind = 
    \begin{cases}
    \frac{1}{\sqrt{\numfeat}}\cos(\rffwt_{\featind}^{T}\onelatent), 1\leq l\leq \numfeat\\
    \frac{1}{\sqrt{\numfeat}}\sin(\rffwt_{\featind-\numfeat}^{T}\onelatent), \numfeat< \featind\leq 2\numfeat
    \end{cases}
\end{align}
The overall kernel is now linear and the computation has complexity $\mathcal{O}(\numtrain^2\numfeat + \numtrain\numweights\numfeat)$.

\subsection{Training Algorithm and its Derivation} 

The forward pass described above, is used to compute the kernel between data points and consequently the predicted GP mean and variance for test data (see \fref{sec:prob}). To train the model we need to find the optimal distribution $\probfunc(\randweights)$ over its parameters in this setting.

Since the latent embedding $\randlatent$ for a data point $\onedata$ is given by $\randlatent=\latfunc_{\randweights}(\onedata)$, $\randweights \sim \probfunc(\randweights)$, we can rewrite \fref{eq:kernij} as 
\begin{align}
\kernel_{ij} &= \mathbb{E}_{\weights,\weights' \sim \probfunc(\randweights)}[\smallkernel(\latfunc_{\weights}(\onedata_i),\latfunc_{\weights'}(\onedata_j))] = \kernel_{ij}[\probfunc] \label{eq:wtkernij}
\end{align}
If Random Fourier Features are used then
\begin{align}
\kernel_{ij} \approx \mathbb{E}_{\weights \sim \probfunc(\randweights)}[\rfffunc_{\rffwtmat}(\latfunc_{\weights}(\onedata_i))]^{T}\mathbb{E}_{\weights' \sim \probfunc(\randweights)}[\rfffunc_{\rffwtmat}(\latfunc_{\weights'}(\onedata_j))] \nonumber
\end{align}
In both cases we can view $\kernel_{ij}$ as a \emph{functional} of $\probfunc(\randweights)$. The overall data likelihood is then also a functional of $\probfunc$ and the negative log likelihood is given by (see \citep{rasmussen2003gaussian})
\begin{align}
    \negll[\probfunc] &= -\log \Pr(\alltarget|\alldata,\probfunc(\randweights)) \label{eq:negll}\\
    &= \frac{1}{2}\alltarget^{T}(\kernmat[\probfunc]+\stdev^{2}\mathbf{I})^{-1}\alltarget + \frac{1}{2}\log (\text{det}(\kernmat[\probfunc] + \stdev^2\mathbf{I}))  \nonumber 
\end{align}
where $\text{det}(\mathbf{A})$ denotes the determinant of a matrix $\mathbf{A}$.

Thus a \emph{maximum likelihood} estimate, $\probfunc^{*}(\randweights)$, of the distribution over model parameters is given by
\begin{align}
    \probfunc^{*}(\randweights) = \argmin_{\probfunc(\randweights) \in \probfuncspace} \negll[\probfunc] \label{eq:DPKL}
\end{align}
where  $\probfuncspace$ is the class of distributions in which we seek $\probfunc^{*}$.

Since the negative log likelihood, $\negll[\probfunc]$, is a \emph{functional} of the distribution $\probfunc$, we can use Functional Gradient Descent to search for distributions in $\probfuncspace$ that minimize its value. However the choice of $\probfuncspace$ is extremely critical to the success of this approach. Previous works in learning GP kernels using neural networks \citep{wilson2016deep,jean2018semi} which seek a point estimate of the model parameters essentially assume that $\probfuncspace = \{\probfunc(\weights)|\probfunc(\weights) = \probfunc(\weights)\delta(\bar{\weights})\}$, i.e., $\probfuncspace$ is the space of Dirac Deltas centered around $\bar{\weights}$. As we show in
\fref{sec:Expts}, these approaches are not very effective in small data settings, possibly due to the restrictive nature of the search space. However, choosing an arbitrary search space might make it challenging to compute $\kernel_{ij}$ as in \fref{eq:wtkernij} due to the difficulty in computing expectations with respect to general high dimensional probability distributions.

Recent works \citep{liu2016stein}, have shown that an effective middle-ground in such settings is to choose $\probfuncspace$ as 
\begin{align}
    \probfuncspace = \{\probfunc(\tfweights)|\tfweights = \weights + \shift(\weights), \weights \sim \probfunc_{0}(\weights), \shift \in \rkhs_\wtkernel\} \label{eq:probfuncspace}
\end{align}
where  $\rkhs_\wtkernel$ is a RKHS given by a kernel $\wtkernel$ between model parameters $\weights$ (note that this is \emph{not} the RKHS $\rkhs_\smallkernel$ into which distributions in the latent space $\latentspace$ are embedded and which is given by the kernel $\smallkernel$). $\probfuncspace$ includes all smooth transformations from the initial distribution $\probfunc_{0}$. Distributions in this set $\probfuncspace$ can closely approximate almost any distribution, particularly those admitting Lipschitz continuous densities \citep{villani2008optimal}. 

Moreover, observe that for any smooth one-to-one transform $\tfweights = \tf(\weights)$, $\weights \sim \probfunc(\randweights)$, the entries of $\kernmat$ under the transformed distribution $\probfunc_{[\tf]}(\tfweights)$ can be written as
\begin{align}
\kernel_{ij} &= \mathbb{E}_{\tfweights,\tfweights' \sim \probfunc_{[\tf]}(\tfweights)}[\smallkernel(\latfunc_{\tfweights}(\onedata_i),\latfunc_{\tfweights'}(\onedata_j))]\\
&= \mathbb{E}_{\weights,\weights' \sim \probfunc(\randweights)}[\smallkernel(\latfunc_{\tf(\weights)}(\onedata_i),\latfunc_{\tf(\weights')}(\onedata_j))]\label{eq:tfwtkernij}.
\end{align}
Since the above holds for infinitesimal shifts $\tfweights = \weights + \shift(\weights)$, a tractable choice of $\probfunc_0$ (For eg. Gaussian), enables efficient approximation of $\kernel_{ij}$ by sample averages with samples $\tfweights_i$, $\tfweights_i = \weights_i + \shift(\weights_i), \weights_i \sim \probfunc_{0}(\weights)$. Thus $\probfuncspace$ is sufficiently large but permits efficient computation.

Moreover $\kernel_{ij}$ in \fref{eq:tfwtkernij} is a \emph{functional} of the transformation $\tf$ (for fixed $\probfunc(\randweights)$) i.e. a functional of the shift $\shift$ in our case. Therefore, the problem of finding $\probfunc^{*}(\randweights)$ in \fref{eq:DPKL} reduces to the problem of finding the optimal shift (given $\probfunc_0(\randweights)$) i.e. 
\begin{align}
    \shift^{*}(\randweights) = \argmin_{\shift(\randweights)} \negll[\shift]. \label{eq:DPKL_shift}
\end{align}
Since $\shift \in \rkhs_\wtkernel$, ($\rkhs_\wtkernel$ is the RKHS for the kernel $\wtkernel$), we can solve \fref{eq:DPKL_shift} via functional gradient descent. The following is an empirical estimate of the functional gradient of the negative log-likelihood at $\shift = 0$. 
\begin{prop}
If we draw $\numweights$ realizations of model parameters $\weights_1,\ldots,\weights_\numweights \sim \probfunc(\weights)$, $\probfunc \in \probfuncspace$, the functional gradient of the negative log likelihood can be approximated as
\begin{align}
    \nabla_{\shift}\negll\mid_{\shift = 0} &\simeq \sum_{l = 1}^{\numweights}\wtkernel(\weights_l,.)\nabla_{\weights_l}\hat{\negll}(\weights_1,\ldots,\weights_\numweights)\label{eq:empfingrad} \\
    \text{where, }\hat{\negll} &= \frac{1}{2}\alltarget^{T}(\hat{\kernmat}+\stdev^{2}\mathbf{I})^{-1}\alltarget + \frac{1}{2}\log(\mbox{det}(\hat{\kernmat} + \stdev^2\mathbf{I})) \nonumber
\end{align}
is the empirical negative log likelihood, $\wtkernel$ is the kernel between model parameters,  $\hat{\kernel}_{ij} = \frac{1}{\numweights^2}\sum_{l,l'=1}^{\numweights}\smallkernel_{ij}(\weights_{l},\weights_{l'})$ are the entries of the empirical kernel matrix $\hat{\kernmat}$.
\end{prop}
To estimate the optimal shift, $\shift^{*}$ (corresponding to the optimal distribution $\probfunc^{*}$) we draw an initial set of samples $\weights_1,\ldots,\weights_\numweights \sim \probfunc_{0}(\weights)$ and iteratively apply the functional gradient descent transformation $\tfweights = \weights - \epsilon\nabla_{\shift}\negll\mid_{\shift=0}$. Since these transformed weights correspond to samples from the transformed distribution $\probfunc_{[\tf]}(\tfweights)$ we only need to evaluate the gradient at $\shift = 0$, which is a crucial advantage since the expression for the gradient at non-zero shifts is much more complicated. The functional gradient in \fref{eq:empfingrad} is a weighted average of individual sample gradients with the kernel between model parameters $\wtkernel$ controlling the effect that different samples have on each other. For $\numweights = 1$ it reduces to gradient descent on negative log likelihood as in Deep Kernel Learning \citep{wilson2016deep}. Algorithm 1 summarises the training procedure for our model, DPKL.

\textbf{Bias in Gradient Estimates.} We note that the gradient estimates in \fref{eq:empfingrad} are biased as the expectation with respect to weights $\weights$ appears inside the matrix inversion in \fref{eq:negll}. While DPKL already outperforms the current state-of-the-art over a wide range of experiments, future work on reducing this bias may give further improvements (as in \citep{belghazi2018mutual}).

\textbf{Connection with SVGD.} While our choice of search space $\probfuncspace$ follows that in Stein's Variational Gradient Descent (SVGD) \citep{liu2016stein} the goal of our work is quite different from theirs. SVGD is a general purpose \emph{Bayesian Inference} algorithm which estimates posterior distributions by minimizing KL divergence w.r.t the true posterior, and has no particular connection to sample efficiency or representation learning, while we seek to improve the quality of learned latent representations and kernels in small data settings by performing \emph{Maximum Likelihood} estimation over the space of probability distributions $\probfuncspace$. The different loss functions also lead to different functional gradient estimates.
\subsection{Extensions}
\label{sec:DPKL_ext}
\textbf{Semi-Supervised Learning (SSDPKL).} While our base model, DPKL, only uses labeled data, it can be augmented with unlabeled data using the regularizer of \citep{jean2018semi} to give a semi-supervised model which we call SSDPKL. Assuming the dataset $\alldata$ has $\numlab$ labeled points $\alldata_L$  (with labels $\alltarget_L$) and $\numunlab$ unlabeled points $\alldata_U$, the new optimization problem is
\begin{align}
    \probfunc^{*}(\randweights) = \argmin_{\probfunc(\randweights) \in \probfuncspace} \frac{1}{\numlab}\negll[\probfunc] + \frac{\alpha}{\numunlab} \sum_{\onedata \in \alldata_U} \stdev^{2}(\onedata). \label{eq:SSDPKL}
\end{align}
Here $\stdev^{2}(\onedata)$ is the GP posterior variance (see \fref{sec:prob}) and $\alpha$ is the regularization parameter. Since \citep{jean2018semi} also restrict $\probfuncspace$ to Dirac Delta functions, our choice of $\probfuncspace$ in \fref{eq:probfuncspace} can generalize this in the same way as it generalizes Deep Kernel Learning. Training involves using functional gradient descent. The following result (illustrated in \fref{fig:SSDPKL}) explains the rationale behind using this regularized loss function in our model.
\begin{prop}
For any unlabeled point, $\onedata \in \alldata_U$, if $\omega_{\onedata}$ is the embedding in the RKHS $\rkhs_\smallkernel$, of $\probfunc(\randlatent|\onedata)$ and $\omega_{\onedata}^{L}$, is its orthogonal projection onto the subspace $\rkhs_\smallkernel^{L} \subseteq \rkhs_\smallkernel$, spanned by RKHS embeddings of $\probfunc(\randlatent|\onedata_i), i=1,\ldots,\numlab, \onedata_i \in \alldata_L$ then, assuming the GP kernel matrix $\kernmat$ is invertible, the posterior variance of $\onedata$ is given by $\stdev^{2}(\onedata) = ||\omega_{\onedata} - \omega_{\onedata}^{L}||_{\rkhs_\smallkernel}^2$.
\end{prop}
Thus minimizing posterior variance encourages the model to learn a mapping that spans distributions in $\latentspace$ corresponding to unlabeled data, which can improve generalization.

\textbf{Classification.} The connection between Gaussian Processes and Bayesian Linear Regression (see \citep{rasmussen2003gaussian}) motivates us to consider Bayesian Logistic Regression (BLR) as our base model for classification. Assuming there are $\numclasses$ classes, the probability of a data point $\onedata_i$ belonging to class $\classind$ under this model is given by
\begin{align}
    \classprob_{i\classind} = \frac{\exp(\classweight_{\classind}^{T}\onedata_i)}{\sum_{\classind'=1}^{\numclasses}\exp(\classweight_{\classind'}^{T}\onedata_i)}, \quad \classweight_\classind \sim \classwtprior(\classweight) \label{eq:blr}
\end{align}
The model can be trained using any Bayesian Inference algorithm to find the optimal posterior distribution over the weights $\classweight_{\classind}$ for each class. When $||\classweight_{\classind}||_2 = ||\onedata_i||_2 = 1$, $\exp(\classweight_{\classind}^{T}\onedata_i) \propto \exp(-\frac{1}{2}||\onedata-\classweight_{\classind}||^{2})$ and \fref{eq:blr} can be rewritten as 
\begin{align}
    \classprob_{i\classind} = \frac{\smallkernel(\onedata_i,\classweight_{\classind})}{\sum_{\classind'=1}^{\numclasses}\smallkernel(\onedata_i,\classweight_{\classind'})}, \quad \classweight_\classind \sim \classwtprior(\classweight) \label{eq:kblr}
\end{align}
where $\smallkernel$ is the SE kernel. Previous works \citep{chen2019closer} have shown that such normalization improves accuracy in small-data (few-shot) classification. This motivates us to replace the high-dimensional data points $\onedata_i$ with low-dimensional \emph{probabilistic embeddings} $\randlatent=\latfunc_{\randweights}(\onedata)$, $\randweights \sim \probfunc(\randweights)$ analogous to DPKL, due to which $\classprob_{i\classind}$ can be written in terms of the corresponding probabilistic kernel as
\begin{align}
    \classprob_{i\classind} = \frac{\mathbb{E}_{\weights \sim \probfunc(\randweights)}[\smallkernel(\latfunc_{\weights}(\onedata_i),\classweight_{\classind})]}{\sum_{\classind'=1}^{\numclasses}\mathbb{E}_{\weights \sim \probfunc(\randweights)}[\smallkernel(\latfunc_{\randweights}(\onedata_i),\classweight_{\classind'})]}, \quad \classweight_\classind \sim \classwtprior(\classweight) \nonumber
\end{align}
To reduce the computational cost the kernel $\smallkernel$ can be approximated by the corresponding Random Fourier Features as in \Fref{sec:overview}. Now in addition to Bayesian Inference for learning the posterior over $\classweight_{\classind}$ we also learn the optimal distribution over model parameters $\probfunc^{*}(\randweights) \in \probfuncspace$ via the Maximum Likelihood estimation approach of Algorithm 1. The negative log likelihood in this case is given by
\begin{align}
    \negll = \sum_{i=1}^{\numtrain}\sum_{\classind=1}^{\numclasses}y_{i\classind}\log\mathbb{E}[\classprob_{i\classind}] \label{eq:crossentropy}
\end{align}
The expectation in \fref{eq:crossentropy} is approximated using samples drawn from the current estimate of the posterior of $\classweight_\classind$. Experiments in \Fref{sec:Expts} show that this model outperforms BLR and is competitive with the state-of-the-art in few-shot classification.
\section{Experimental Results}
\label{sec:Expts}
\begin{figure*}[t]
  \centering
   \subfloat[CTSlice ($\highdim = 384$)]{\includegraphics[width=0.33\linewidth]{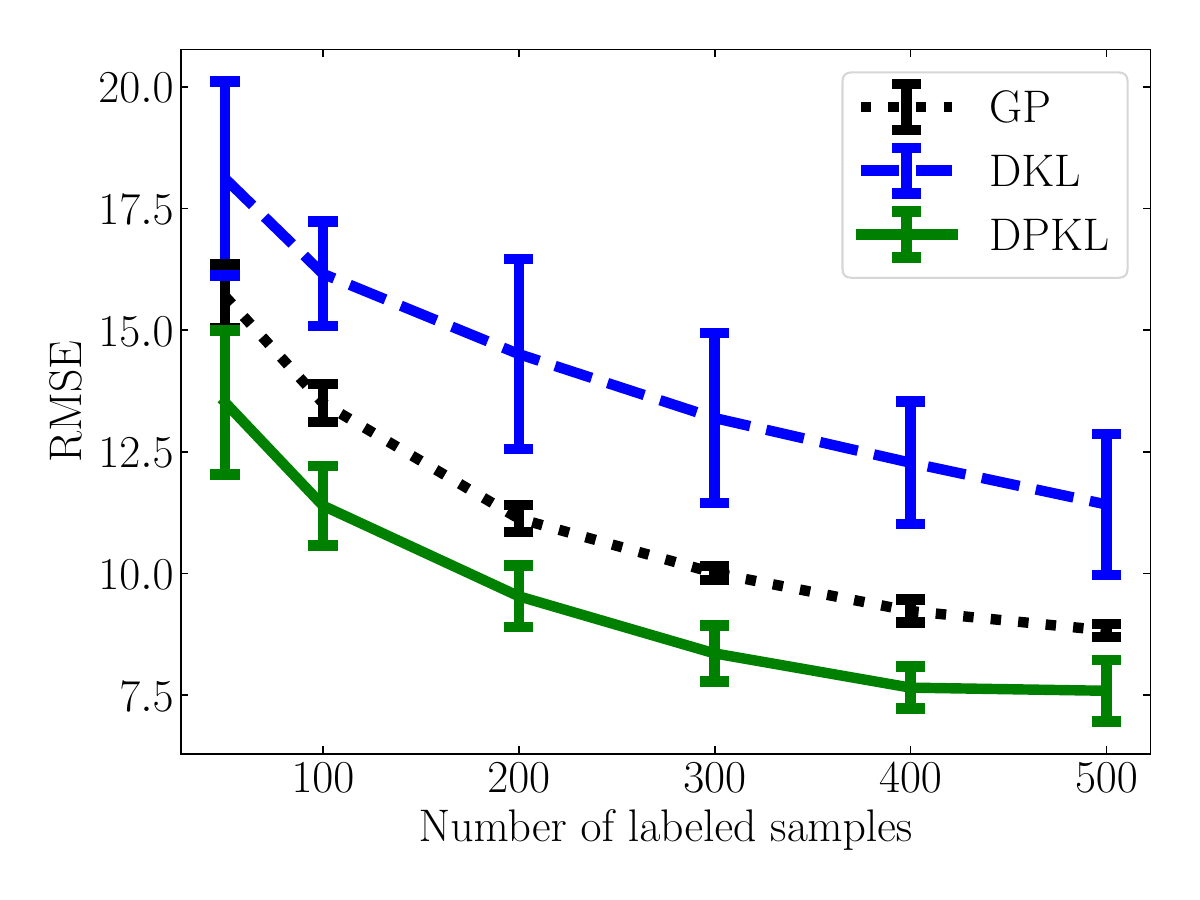}}
   \subfloat[Electric ($\highdim = 6$)]{\includegraphics[width=0.33\linewidth]{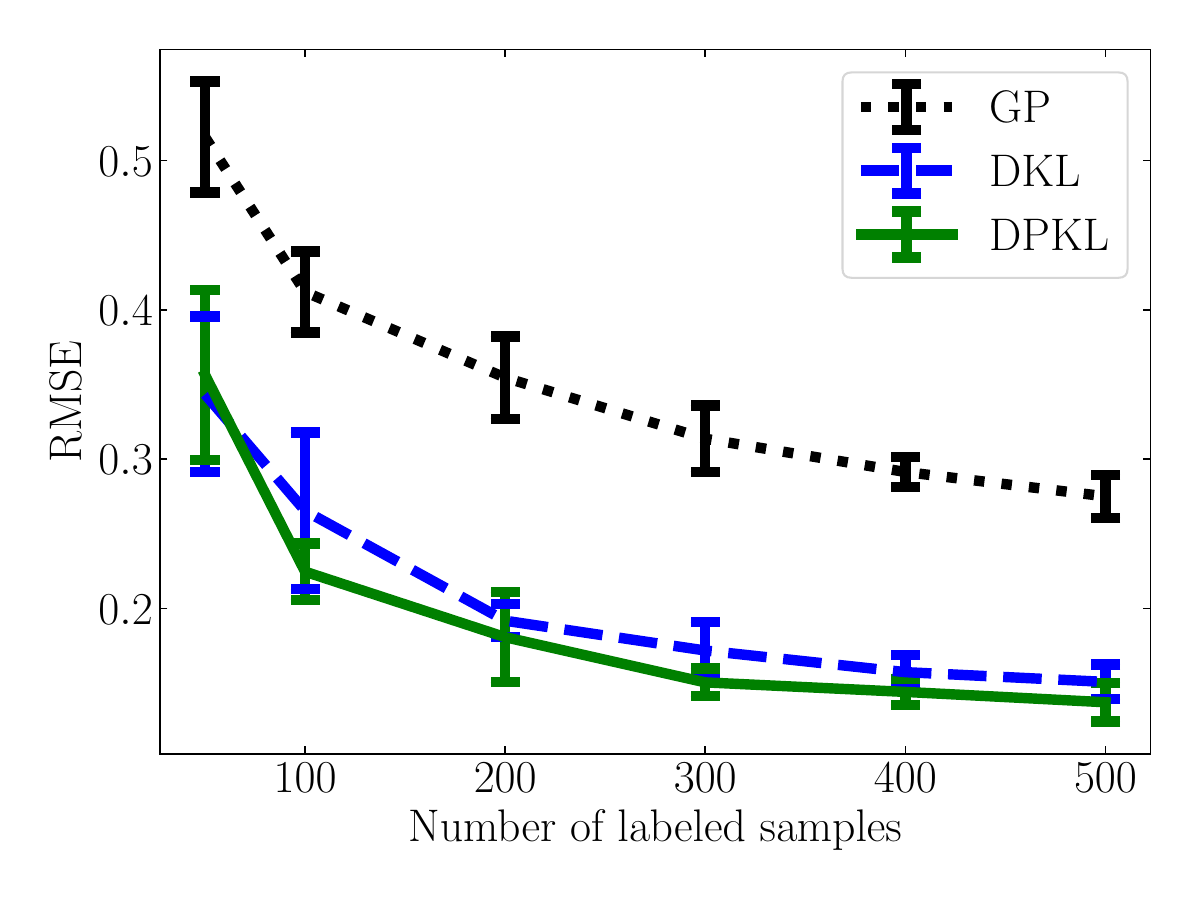}}
   \subfloat[Parkinsons ($\highdim = 20$)]{\includegraphics[width=0.33\linewidth]{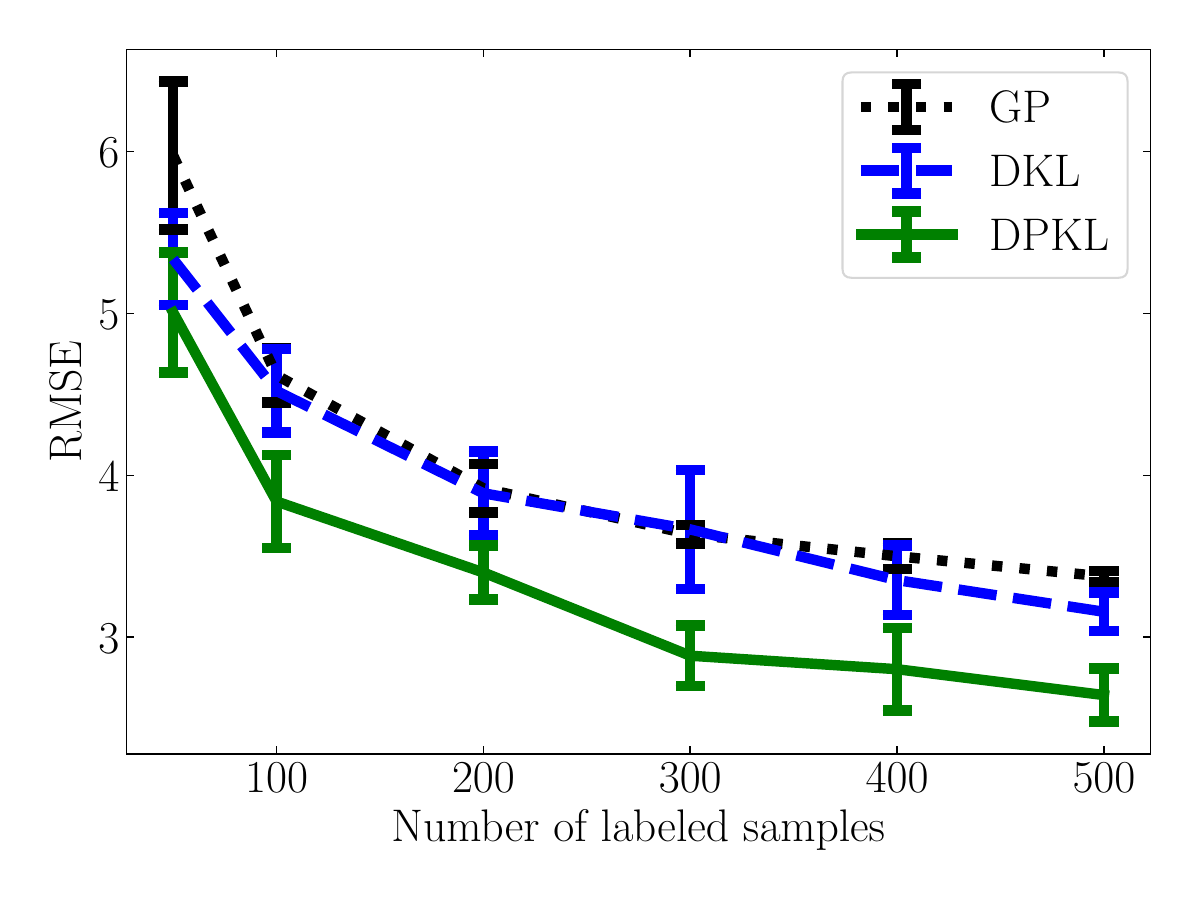}}\\
    \subfloat[Skillcraft ($\highdim = 18$)]{\includegraphics[width=0.33\linewidth]{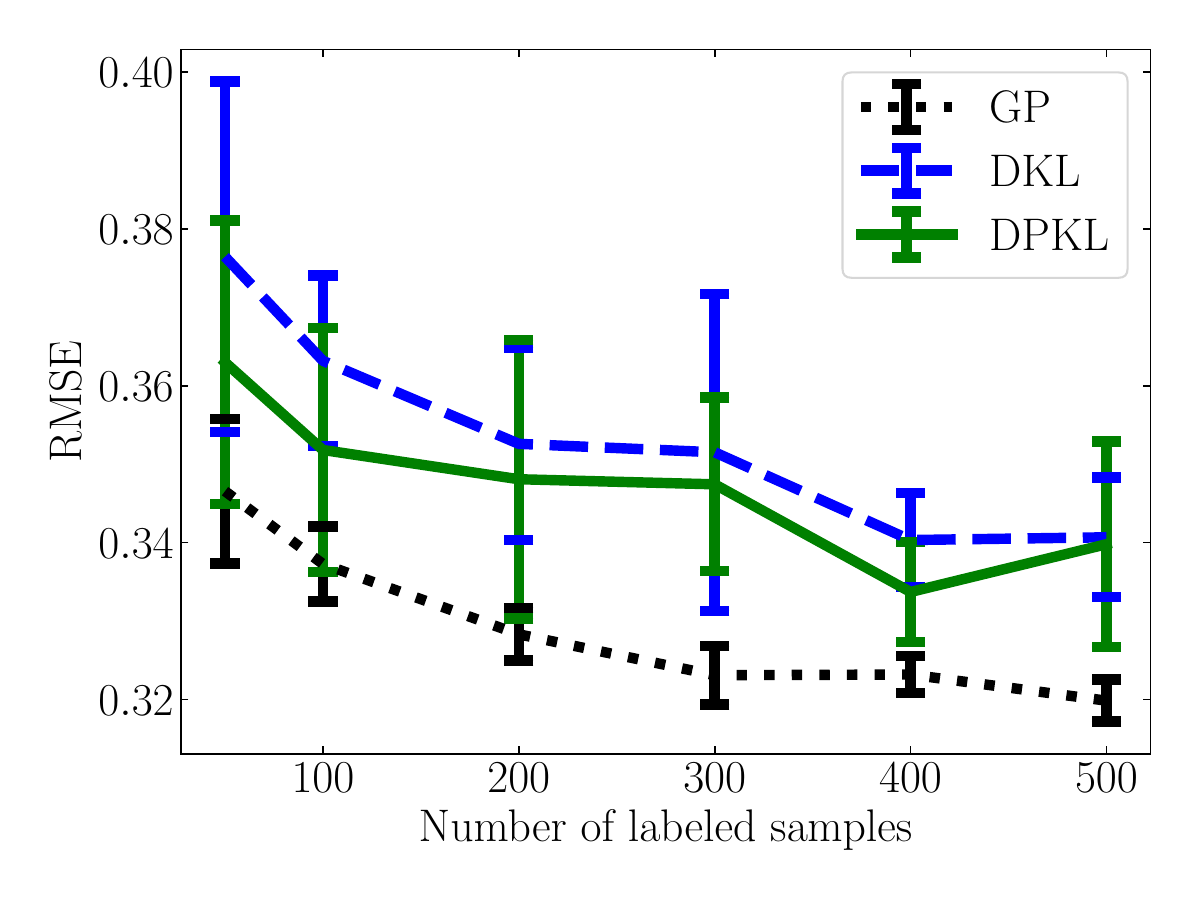}}
   \subfloat[Buzz ($\highdim = 77$)]{\includegraphics[width=0.33\linewidth]{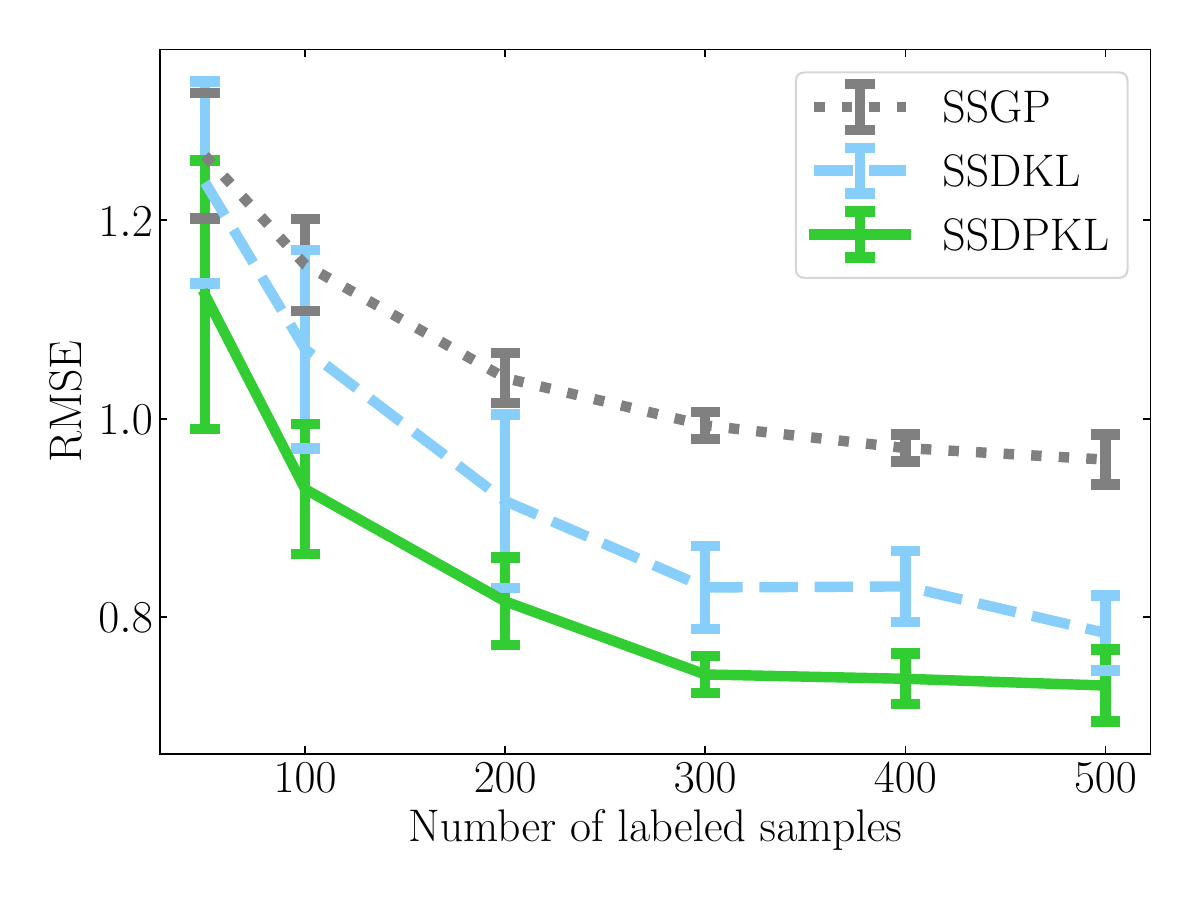}}
   \subfloat[Elevators ($\highdim = 18$)]{\includegraphics[width=0.33\linewidth]{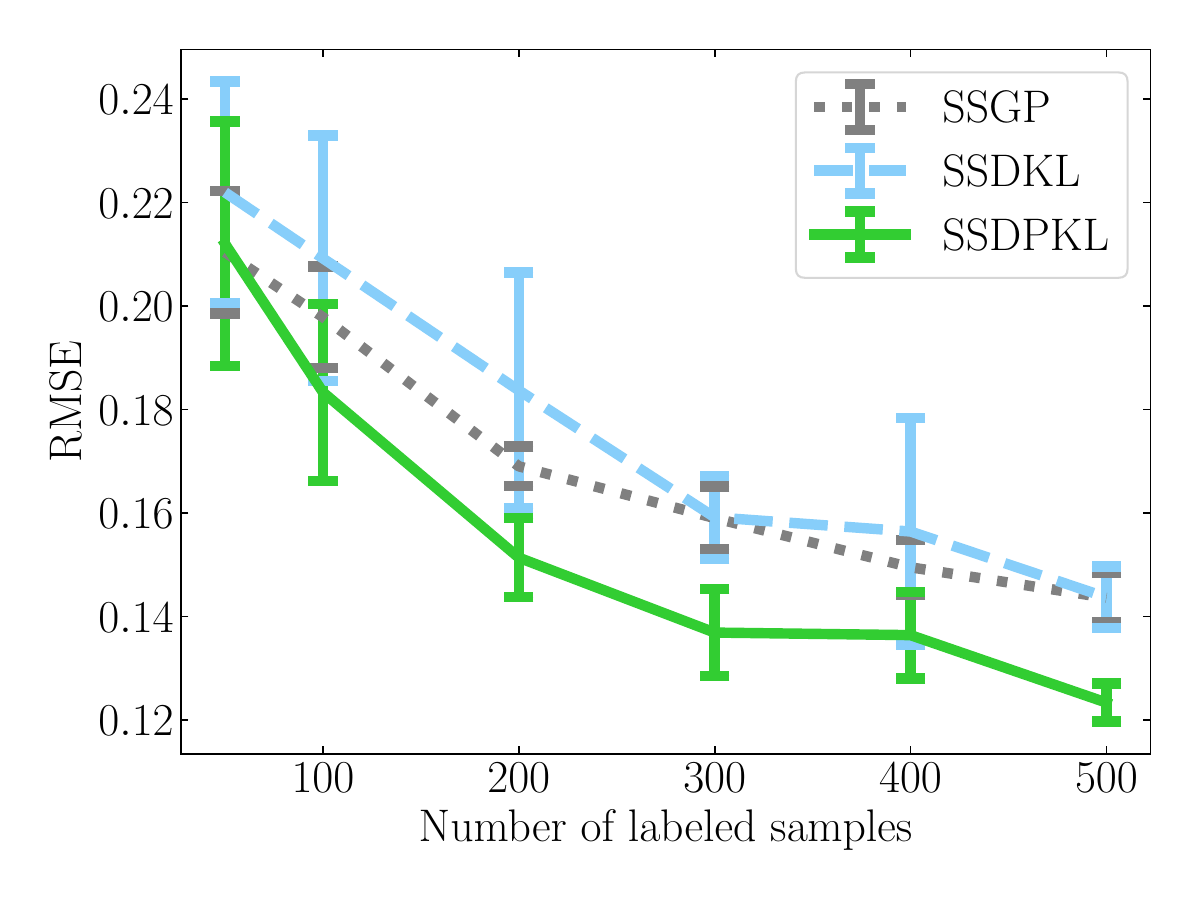}}
   \caption{RMSE Results for regression on 6 UCI Datasets with $\numtrain = \{50, 100, 200, 300, 400, 500\}$ labeled samples. Here show results for DPKL v/s DKL \citep{wilson2016deep} on 4 datasets and the semi-supervised models SSDPKL v/s SSDKL on the other 2 datasets. The clear superiority of DPKL and SSDPKL over their deterministic counterparts (DKL and SSDKL) confirms the efficacy of our approach. See Appendix B for the remaining results.}
  \label{fig:UCI}
\end{figure*}
We apply our model DPKL, to various classification and regression tasks in the small data regime. All models are implemented in TensorFlow \citep{abadi2016tensorflow}. We use the SE kernel everywhere as the kernel $\kappa$ between DPKL model parameters $\weights$ with bandwidth chosen according to the median heuristic described in \citep{liu2016stein}. The Code for the experiments is available at \href{https://github.com/ankurmallick/DPKL}{https://github.com/ankurmallick/DPKL}. 

\textbf{UCI Regression.} We use DPKL, and its semi-supervised variant, SSDPKL for regression on 6 datasets of varying dimensionality from the UCI repository \citep{lichman2013uci} as shown in \Fref{fig:UCI}. We compare our models to DKL \citep{wilson2016deep} and SSDKL \citep{jean2018semi} (both of which use deterministic neural networks to map data into the latent space), and GP and Semi-Superivsed GP (SSGP) (regularized as in \Fref{eq:SSDPKL}). The GP and SSGP models use a Squared Exponential (SE) kernel directly on the data i.e. $\smallkernel(\onedata_i,\onedata_j) = \exp(-\frac{1}{2}\sum_{l}\frac{1}{h_{l}^{2}}(\onedata_{il} - \onedata_{jl})^2)$. The DKL and SSDKL models embed the data into a low-dimensional latent space before using a SE kernel i.e. $\smallkernel(\onedata_i,\onedata_j) = \exp(-\frac{1}{2}\norm{\latfunc_{\weights}(\onedata_i) - \latfunc_{\weights}(\onedata_j)})$. The DPKL and SSDPKL models embed the data into a \emph{probability distribution} in the latent space (as described in \Fref{sec:DPKL}) and then use the SE kernel between distributions as given in \Fref{eq:kernij_rbf}. We used the same neural network architecture as \citep{jean2018semi} ($\highdim - 100 - 50 - 50 - \lowdim$) for mapping data points to the latent space in the DKL and DPKL models. Here $\highdim$ is the dimensionality of the datasets and $\lowdim = 2$ is the dimensionality of the latent embedding $\onelatent \in \latentspace$. In DPKL and SSDPKL, we used $\numweights = 10$ samples from this architecture to represent the distribution over model parameters. The lengthscales for the GP SE kernel, and the neural network weights in DKL are optimized using gradient descent on the negative log-likelihood while the \emph{distribution} over neural network weights in DPKL is optimized using functional gradient descent (Algorithm 1). For each dataset we use $\numtrain = \{50, 100, 200, 300, 400, 500\}$ labeled examples. The semi-supervised models -- SSDPKL, SSDKL and SSGP, additionally have access to (at most) $10000$ unlabeled examples from the corresponding datasets while the supervised models, DPKL, DKL, and GP, do not use any unlabeled data. More details on training procedure are in Appendix C.

\Fref{fig:UCI} shows RMSE for supervised models on 4 datasets and for semi-supervised models on the other 2 datasets. Clearly DPKL (SSDPKL) improves over DKL (SSDKL) across \emph{all} datasets. While GP does a bit better than DPKL on one of the lower dimensional datasets (Skillcraft), we note that DPKL is significantly better than GP on all other datasets, including the high dimensional datasets CTSlice ($D = 384$) and Buzz ($D = 77$), thus clearly illustrating the advantages of our approach in high dimensional settings.

\begin{figure*}[t]
  \centering
   \subfloat[CTSlice ($\highdim = 384$)]{\includegraphics[width=0.33\linewidth]{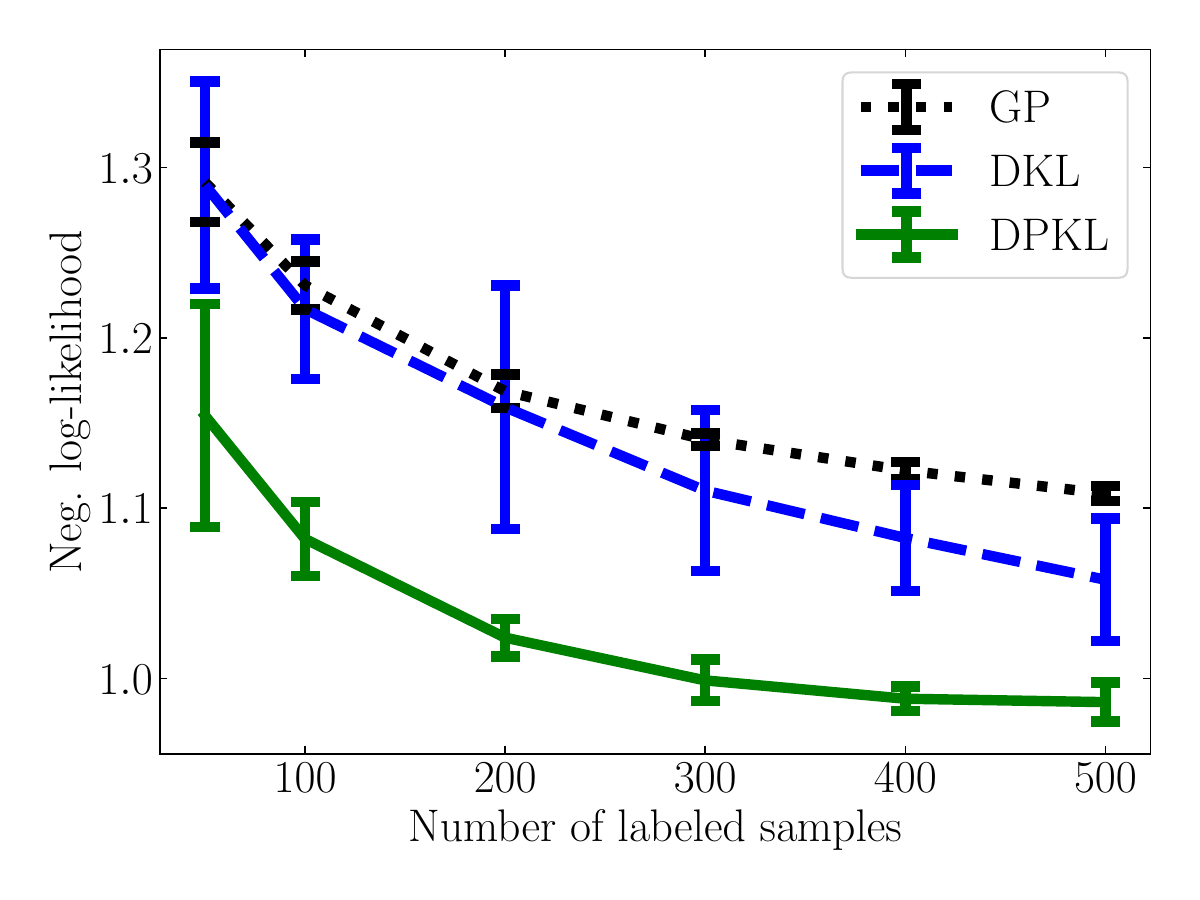}}
   \subfloat[Electric ($\highdim = 6$)]{\includegraphics[width=0.33\linewidth]{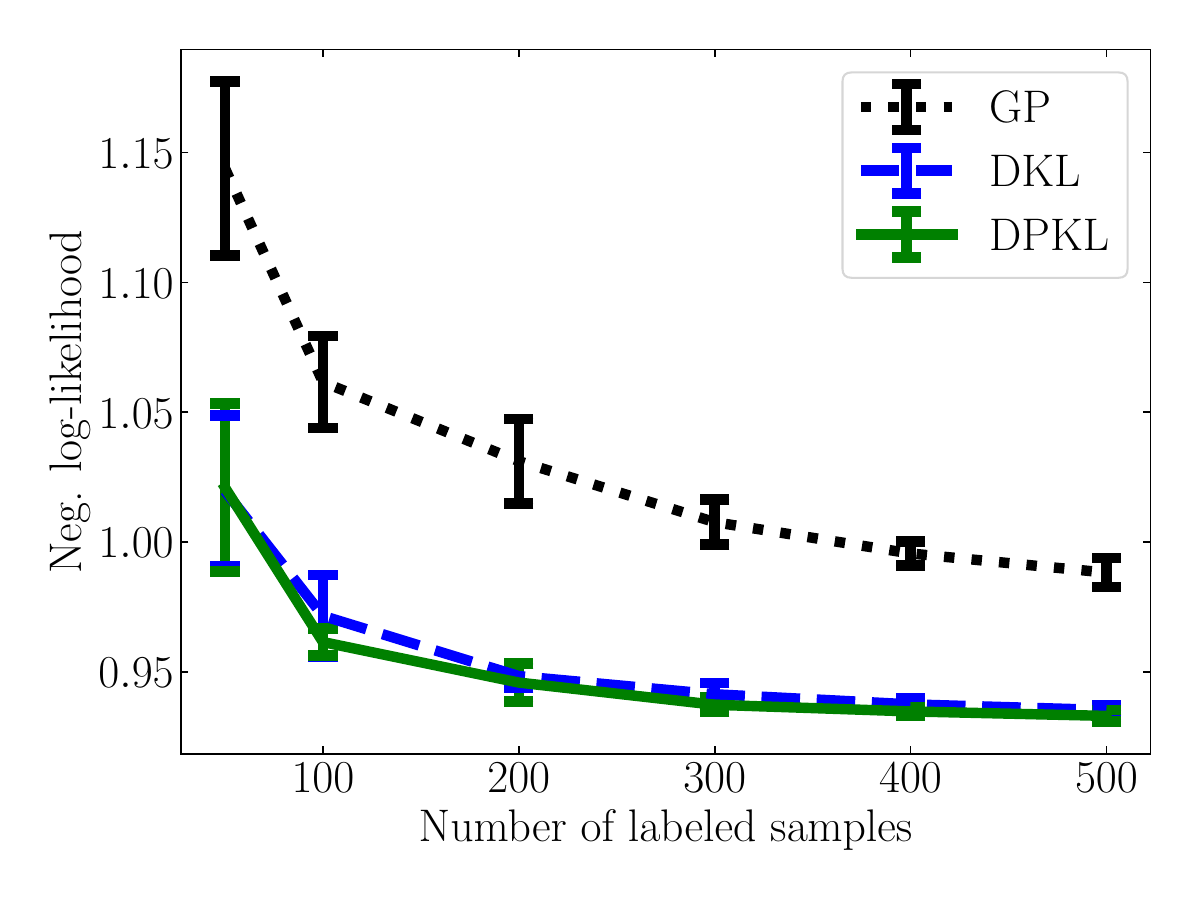}}
   \subfloat[Parkinsons ($\highdim = 20$)]{\includegraphics[width=0.33\linewidth]{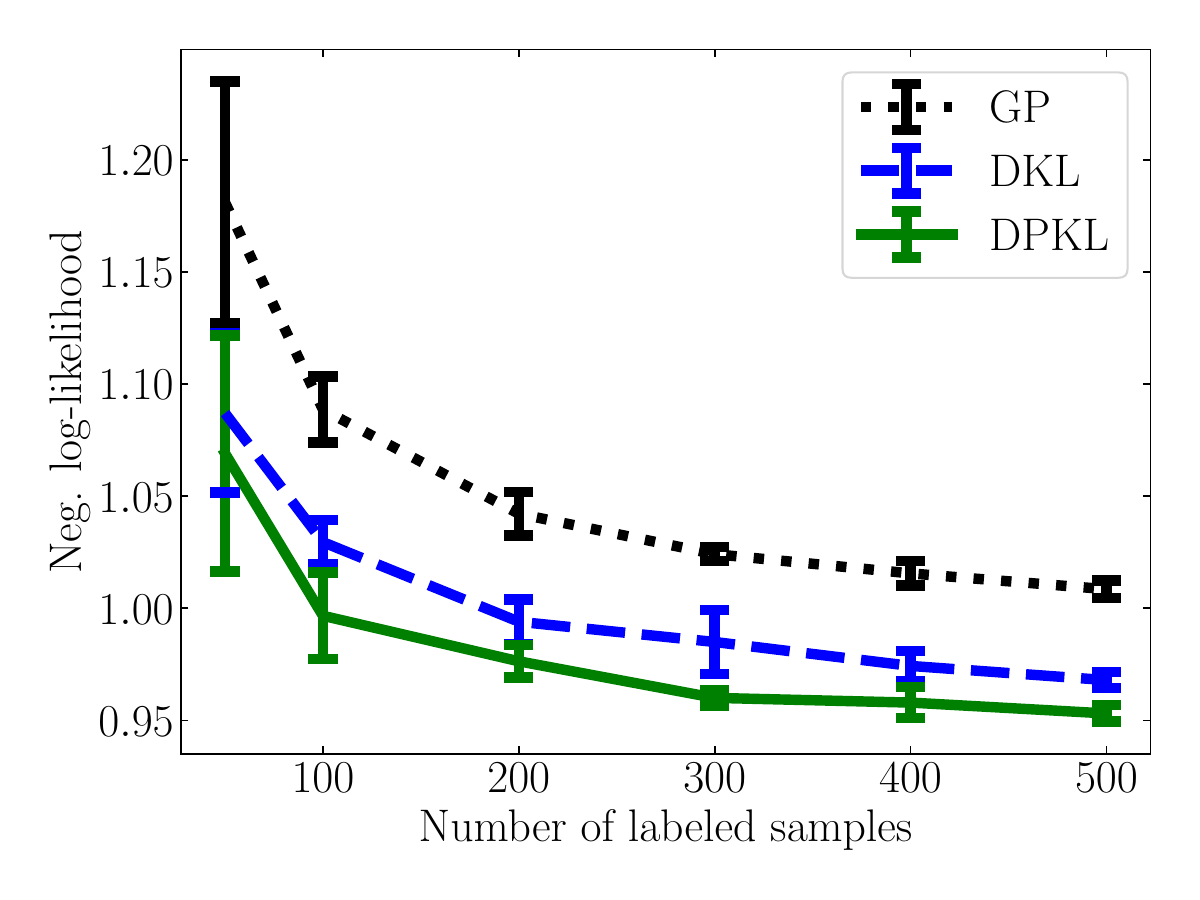}}\\
    \subfloat[Skillcraft ($\highdim = 18$)]{\includegraphics[width=0.33\linewidth]{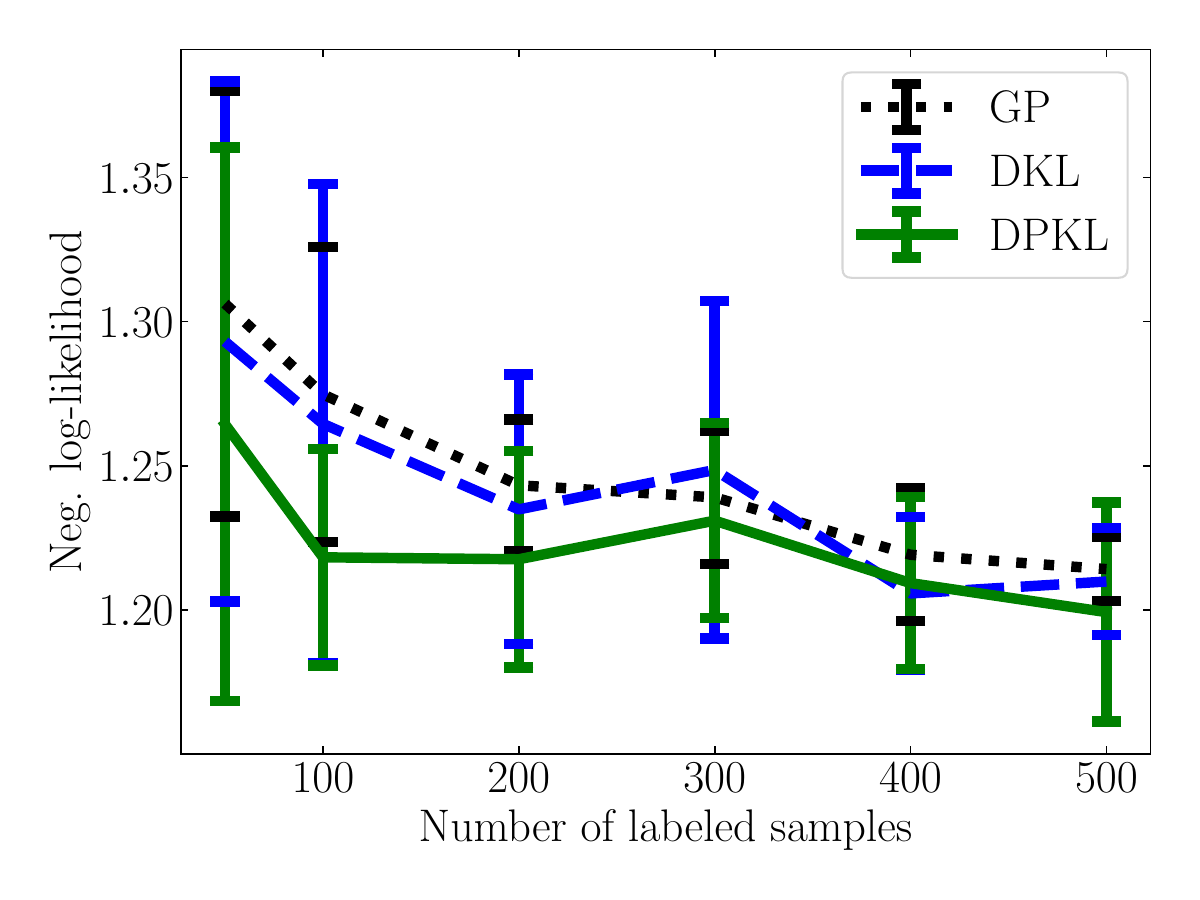}}
   \subfloat[Buzz ($\highdim = 77$)]{\includegraphics[width=0.33\linewidth]{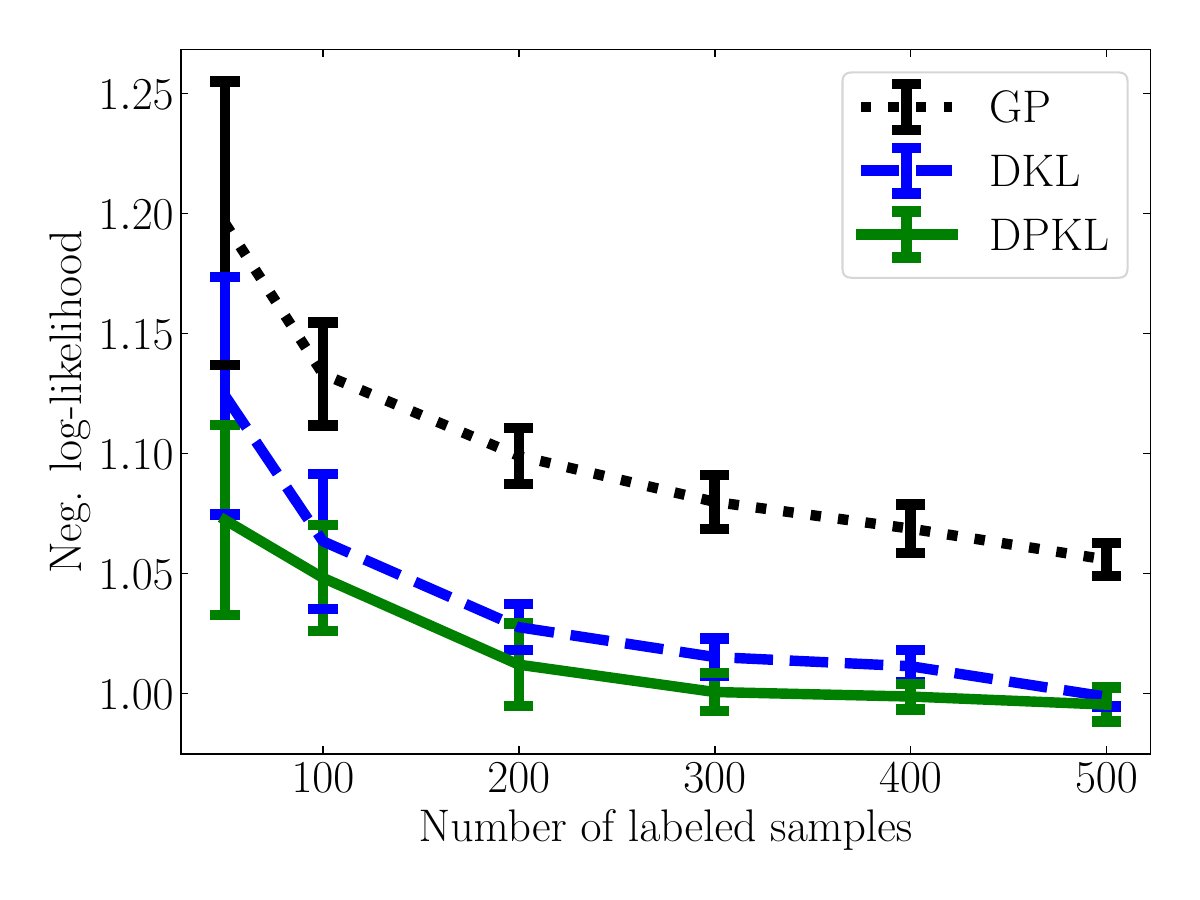}}
   \subfloat[Elevators ($\highdim = 18$)]{\includegraphics[width=0.33\linewidth]{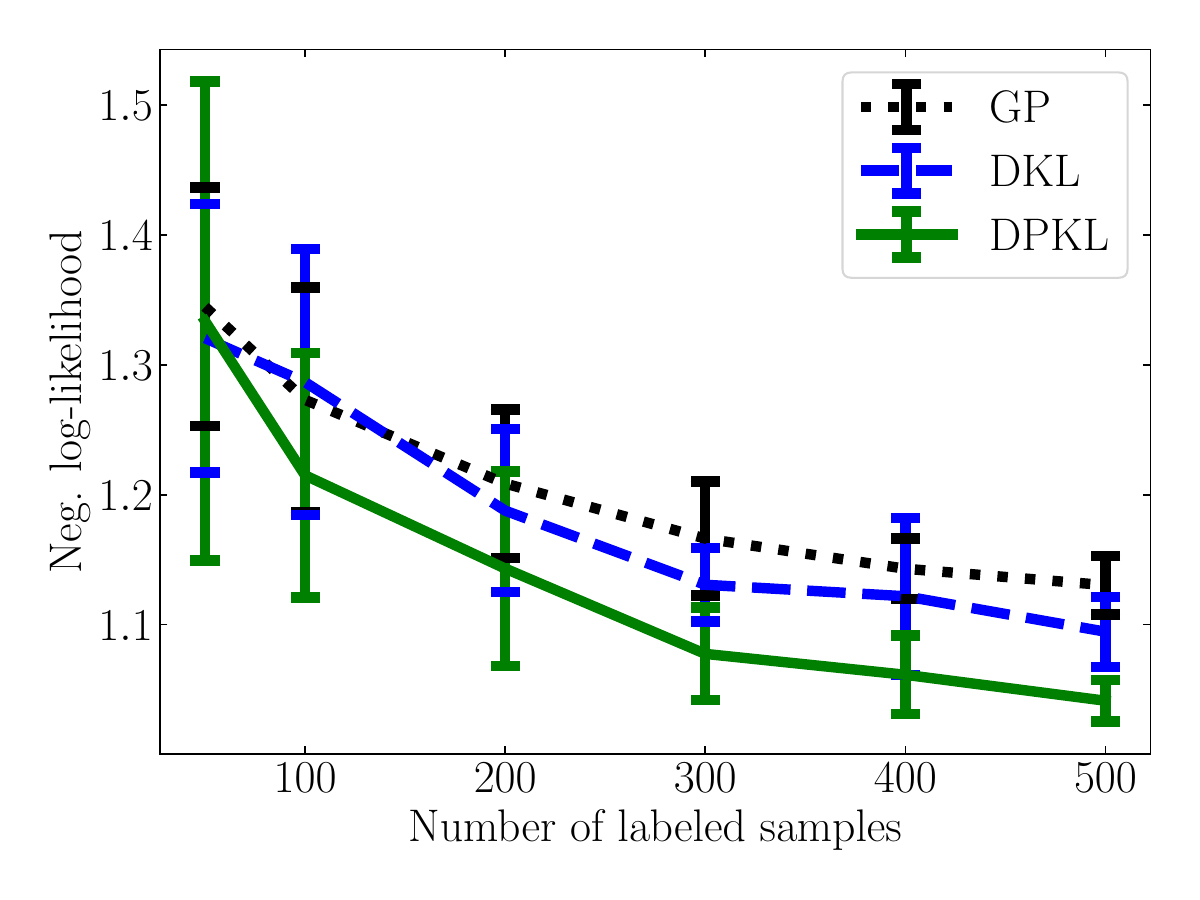}}
   \caption{Uncertainty Quantification (average negative log-likelihood of test data) results for regression on 6 UCI Datasets with $\numtrain = \{50, 100, 200, 300, 400, 500\}$ labeled samples.  for DPKL, DKL and GP. Lower negative log-likelihood indicates better uncertainty quantification. DPKL generally has lower negative log-likelihood than GP and DKL.}
  \label{fig:UCI_UQ}
\end{figure*}

An added benefit of our approach is better uncertainty quantification. Since all Gaussian Process models output a variance in addition to the mean prediction we use the average negative log likelihood of test data as in \citep{lakshminarayanan2017simple} to quantify the uncertainty of model predictions. A lower negative log likelihood implies that the model fits the test data better. The results in \Fref{fig:UCI_UQ} shows that by projecting high dimensional data to low dimensional probability distributions before making predictions, DPKL consistently outperforms DKL and GP in this metric across all datasets.

\textbf{Few-shot Classification.} We apply DPKL to the 5-shot classification task described in \citep{chen2019closer}. Here a pre-trained base model with a deterministic feature extractor (4-layer convolution backbone) attached to a classifier (any suitable classifier), is fine-tuned on a small number of examples (5 per class) of classes that were \emph{unseen} during training. Accuracy is measured on a test dataset of examples from the \emph{unseen classes}. We keep the backbone (feature extractor) fixed and change the classifier in our experiments. We compare the DPKL Classifier (described in \Fref{sec:DPKL_ext}) to the Baseline++ with 1-NN classifier described in Appendix A4 of \citep{chen2019closer} since it is the deterministic version (deterministic embedding, deterministic logistic regression) of the DPKL classification approach, and also to a classifier consisting of a deterministic embedding layer followed by Bayesian Logistic Regression (similar to \citep{snoek2015scalable}) which we call DBLR. Results in Table 1 show that DPKL outperforms Baseline++ and DBLR for 5-shot classification on both the CUB and mini-Imagenet datasets which were considered in \citep{chen2019closer}. DPKL outperforming DBLR indicates that improvements are not obtained just by replacing deterministic logistic regression with Bayesian logistic regression but by \emph{probabilistic embeddings} learned in DPKL. More details on training procedure are in Appendix C and additional classification experiments are presented in Appendix B.
\begin{table}[t]
\begin{center}
\begin{tabular}{ |c c c| } 
 \hline
 \textbf{Method} & \textbf{CUB} & \textbf{mini-Imagenet}\\
 \hline
 \textbf{Baseline++} & $68.94 \pm 0.74$ & $61.93 \pm 0.65$ \\ 
 \textbf{DBLR} & $68.48 \pm 0.71$ & $60.12 \pm 0.63$ \\ 
 \textbf{DPKL (Ours)} & $\mathbf{69.11 \pm 0.72}$ & $\mathbf{64.96 \pm 0.62}$ \\ 
  \hline
\end{tabular}
\end{center}
\caption{DPKL has higher accuracy than the Baseline++ method of \citep{chen2019closer} as well as DBLR for 5-shot classification on both CUB and mini-Imagenet datasets.}
\label{tbl:few_shot}
\end{table} 
\section{Conclusion}
We propose a new approach for small data learning that maps high dimensional data to low dimensional probability distributions and then performs regression/classification on these distributions. The distribution over model parameters is learned via functional gradient descent. Our model outperforms several baselines in GP regression and few-shot classification while learning a meaningful representation of the data and accurately quantifying uncertainties on test data. In future we plan to theoretically analyze the convergence of our approach, seek potential improvements through optimal kernel selection and unbiased gradient estimation, and apply our model to areas like Bayesian Optimization \citep{brochu2010tutorial} where GPs have been successful in the past.

\section{Acknowledgments}
This work was supported in part by Lawrence Livermore National Laboratory under Contract DE-AC52-07NA27344 and LLNL-LDRD Program Project No. 19-SI-001, and the Army Research Office under the ARO Grant W911NF-16-1-0441. 

\begin{appendix}
\section{Proof of Theoretical Results}
\label{sec:proofs_app}
\subsection{Proof of Proposition 1}

Recall that the entries of the kernel matrix $\kernel_{ij}$ under the transformation $\tfweights = \tf(\weights) = \weights + \shift(\weights)$ are given by
\begin{align}
    \kernel_{ij} &= \mathbb{E}_{\tfweights,\tfweights' \sim \probfunc_{[\tf]}(\randtfweights)}[\smallkernel(\latfunc_{\tfweights}(\onedata_i),\latfunc_{\tfweights'}(\onedata_j))] \\
    &= \mathbb{E}_{\weights,\weights' \sim \probfunc(\randweights)}[\smallkernel(\latfunc_{\tf(\weights)}(\onedata_i),\latfunc_{\tf(\weights')}(\onedata_j))]\label{eq:tfwtkernij_app}
\end{align}

Defining $\smallkernel_{ij}(\weights,\weights') = \smallkernel(\latfunc_{\weights}(\onedata_i),\latfunc_{\weights'}(\onedata_j))$ we have
\begin{align}
    \kernel_{ij} = \mathbb{E}_{\weights,\weights' \sim \probfunc(\randweights)}[\smallkernel_{ij}(\weights + \shift(\weights),\weights' + \shift(\weights'))]\label{eq:shiftwtkernij}
\end{align}

Assuming that the distributions $\probfunc(\randweights)$ and shifts $\shift(\randweights)$ are functions in a RKHS $\mathcal{H}$ given by the kernel $\wtkernel$ ($\wtkernel$, $\smallkernel$, and $\kernel$ are all different), we have (from the definition of functional gradient $\nabla_{\shift}\kernel_{ij}[\shift]$),
\begin{align}
    \kernel_{ij}[\shift + \epsilon\altshift] = \kernel_{ij}[\shift] + \epsilon<\nabla_{s}\kernel_{ij}[\shift],\altshift>_{\rkhs} + \mathcal{O}(\epsilon^2)
\end{align}

Thus we need to compute the difference $\kernel_{ij}[\shift + \epsilon\altshift] - \kernel_{ij}[\shift]$ which, from \Fref{eq:shiftwtkernij} is given by
\begin{align}
     \begin{split}
        \kernel_{ij}[\shift + \epsilon\altshift] - \kernel_{ij}[\shift] = \mathbb{E}_{\probfunc}[\smallkernel_{ij}(\weights + \shift(\weights) + \epsilon\altshift(\weights),\weights' + \shift(\weights') + \epsilon\altshift(\weights'))] \\
        - \mathbb{E}_\probfunc[\smallkernel_{ij}(\weights + \shift(\weights),\weights' + \shift(\weights'))]\\
    \end{split}
\end{align}

We use $\mathbb{E}_{\probfunc}$ to denote the expectation when $\weights,\weights' \sim \probfunc(\randweights)$. The above equation can be rewritten as $\kernel_{ij}[\shift + \epsilon\altshift] - \kernel_{ij}[\shift] = V_1 + V_2$ where
\begin{align}
    V_1 &= \mathbb{E}_{\probfunc}[\smallkernel_{ij}(\weights + \shift(\weights) + \epsilon\altshift(\weights),\weights' + \shift(\weights') + \epsilon\altshift(\weights'))] - \mathbb{E}_{\probfunc}[\smallkernel_{ij}(\weights + \shift(\weights) ,\weights' + \shift(\weights') + \epsilon\altshift(\weights'))]\\
    & = \epsilon\mathbb{E}_{\probfunc}[\nabla_{\weights}\smallkernel_{ij}(\weights +\shift(\weights),\weights' + \shift(\weights') + \epsilon\altshift(\weights'))\altshift(\weights))] +\mathcal{O}(\epsilon^2)\\
    \begin{split}
    &=\epsilon\mathbb{E}_{\probfunc}[(\nabla_{\weights}\smallkernel_{ij}(\weights +\shift(\weights),\weights' + \shift(\weights') + \epsilon\altshift(\weights')) \\
    &- \nabla_{\weights}\smallkernel_{ij}(\weights +\shift(\weights),\weights' + \shift(\weights')) + \nabla_{\weights}\smallkernel_{ij}(\weights +\shift(\weights),\weights' + \shift(\weights')))\altshift(\weights))] +\mathcal{O}(\epsilon^2)
    \end{split}\\
    &=\epsilon\mathbb{E}_{\probfunc}[\nabla_{\weights}\smallkernel_{ij}(\weights +\shift(\weights),\weights' + \shift(\weights')))\altshift(\weights))] + \mathcal{O}(\epsilon^2)\\
    &=\epsilon<\mathbb{E}_{\probfunc}[\nabla_{\weights}\smallkernel_{ij}(\weights +\shift(\weights),\weights' + \shift(\weights')))\wtkernel(\weights,.))],\altshift>_{\rkhs} + \mathcal{O}(\epsilon^2)
\end{align}
where the last line follows from the RKHS property.
Similarly,
\begin{align}
    V_2 &=\mathbb{E}_{\probfunc}[\smallkernel_{ij}(\weights + \shift(\weights) ,\weights' + \shift(\weights') + \epsilon\altshift(\weights'))] - \mathbb{E}_\probfunc[\smallkernel_{ij}(\weights + \shift(\weights),\weights' + \shift(\weights'))]\\
    &=\epsilon<\mathbb{E}_{\probfunc}[\nabla_{\weights'}\smallkernel_{ij}(\weights +\shift(\weights),\weights' + \shift(\weights')))\wtkernel(\weights',.))],\altshift>_{\rkhs} + \mathcal{O}(\epsilon^2)
\end{align}

Since we transform the weights $\weights$ after every iteration, therefore we only ever need to compute the gradient at $\shift(\weights) = 0$. Thus, finally, we have the expression
\begin{align}
    \nabla_{\shift}\kernel_{ij}[\shift]\mid_{\shift = 0} = \mathbb{E}_{\probfunc}[\nabla_{\weights}\smallkernel_{ij}(\weights,\weights' )\wtkernel(\weights,.) + \nabla_{\weights'}\smallkernel_{ij}(\weights,\weights')\wtkernel(\weights',.)]\label{eq:fingrad_app}
\end{align}

If we draw $\numweights$ samples of model parameters $\weights_1,\ldots,\weights_\numweights \sim \probfunc(\weights)$, the empirical estimate of $ \nabla_{\shift}\kernel_{ij}[\shift]\mid_{\shift = 0}$ given by replacing expectations with sample averages is given by
\begin{align}
    \nabla_{\shift}\kernel_{ij}[\shift]\mid_{\shift = 0} \simeq \frac{1}{\numweights^2}\sum_{l,l' = 1}^{\numweights}[\nabla_{\weights_l}\smallkernel_{ij}(\weights_{l},\weights_{l'}))\wtkernel(\weights_l,.)) + \nabla_{\weights_{l'}}\smallkernel_{ij}(\weights_{l},\weights_{l'}))\wtkernel(\weights_{l'},.))]\label{eq:empfingrad_app}
\end{align}
Without loss of generality consider all the terms in the above expression that contain the gradient with respect to $\weights_1$ and let us call that part of the summation $T_1$. Therefore
\begin{align}
    T_1 = \frac{1}{\numweights^2}\sum_{l' = 1}^{\numweights}\nabla_{\weights_1}\smallkernel_{ij}(\weights_{1},\weights_{l'}))\wtkernel(\weights_1,.)) + \frac{1}{\numweights^2}\sum_{l = 1}^{\numweights}\nabla_{\weights_1}\smallkernel_{ij}(\weights_{l},\weights_{1}))\wtkernel(\weights_1,.))\label{eq:empfingrad_w1}
\end{align}
Recall the expression for the empirical estimate of the entries of the kernel matrix $\kernel_{ij}$
\begin{align}
    \hat{\kernel}_{ij} \simeq \frac{1}{\numweights^2}\sum_{l,l' = 1}^{\numweights}\smallkernel_{ij}(\weights_{l},\weights_{l'})
\end{align}
Differentiating both sides with respect to $\weights_1$,
\begin{align}
    \nabla_{\weights_1}\hat{\kernel}_{ij} \simeq \frac{1}{\numweights^2}\sum_{l' = 1}^{\numweights}\nabla_{\weights_1}\smallkernel_{ij}(\weights_{1},\weights_{l'}) + \frac{1}{\numweights^2}\sum_{l = 1}^{\numweights}\nabla_{\weights_1}\smallkernel_{ij}(\weights_{l},\weights_{1}) \label{eq:empkerngrad_w1}
\end{align}
Note that the term $\nabla_{\weights_1}\smallkernel_{ij}(\weights_{1},\weights_{1})$ occurs in both summmations. This is because $\nabla_{\weights_1}\smallkernel_{ij}(\mathbf{u},\mathbf{v}) = \nabla_{\mathbf{u}}\smallkernel_{ij}(\mathbf{u},\mathbf{v})\nabla_{\weights_1}\mathbf{u} + \nabla_{\mathbf{v}}\smallkernel_{ij}(\mathbf{u},\mathbf{v})\nabla_{\weights_1}\mathbf{v} = \nabla_{\weights_1}\smallkernel_{ij}(\weights_{1},\weights_{1}) + \nabla_{\weights_1}\smallkernel_{ij}(\weights_{1},\weights_{1})$ when $\mathbf{u} = \mathbf{v} = \mathbf{\weights_1}$ ($\mathbf{u}, \mathbf{v}, \mathbf{\weights_1}$ are all variable)
.

Substituting \Fref{eq:empkerngrad_w1} in \Fref{eq:empfingrad_w1}
\begin{align}
    T_1 = \wtkernel(\weights_1,.)\nabla_{\weights_1}\hat{\kernel}_{ij}
\end{align}
We can apply the same argument to simplify the terms in \Fref{eq:empfingrad_app} that contain gradients with respect to other weights $\weights_2,\ldots,\weights_\numweights$ in the same fashion. Therefore,
\begin{align}
    \nabla_{\shift}\kernel_{ij}[\shift]\mid_{\shift = 0} &\simeq \frac{1}{\numweights^2}\sum_{l,l' = 1}^{\numweights}[\nabla_{\weights_l}\smallkernel_{ij}(\weights_{l},\weights_{l'}))\wtkernel(\weights_l,.)) + \nabla_{\weights_{l'}}\smallkernel_{ij}(\weights_{l},\weights_{l'}))\wtkernel(\weights_{l'},.))]\\
    &= \sum_{l = 1}^{\numweights}\wtkernel(\weights_l,.)\nabla_{\weights_l}\hat{\kernel}_{ij}
\end{align}

From the chain rule for functional gradient descent we have $\nabla_{\shift}\negll = \sum_{i,j}\frac{\partial \negll}{\partial \kernel_{ij}}\nabla_{\shift}\kernel_{ij}[\shift]$ and the corresponding empirical estimate $\nabla_{\shift}\negll\mid_{\shift = 0} \simeq \sum_{i,j}\frac{\partial \hat{\negll}}{\partial \hat{\kernel}_{ij}}\nabla_{\shift}(\sum_{l = 1}^{\numweights}\wtkernel(\weights_l,.)\nabla_{\weights_l}\hat{\kernel}_{ij})$. Switching the order of the summations gives
\begin{align}
    \nabla_{\shift}\negll\mid_{\shift = 0} \simeq \sum_{l = 1}^{\numweights}\wtkernel(\weights_l,.)\nabla_{\weights_l}\hat{\negll}(\weights_1,\ldots,\weights_\numweights)
\end{align}

\subsection{Proof of Proposition 2}

From the theory of kernel embeddings of probability distributions \citep{muandet2017kernel}, we know that for any data point $\onedata$ the embedding of the distribution $p(\randlatent|\onedata)$ in the RKHS $\rkhs_\smallkernel$ is given by
\begin{align}
    \rkhsvect_\onedata = \int \smallkernel(\onelatent,.)\probfunc(\onelatent|\onedata)d\onelatent
\end{align}
where $\smallkernel$ is the kernel corresponding to the RKHS $\rkhs_\smallkernel$. Under our model $\randlatent = \latfunc_{\randweights}(\onedata)$ where $\randweights \sim \probfunc(\randweights)$. Therefore
\begin{align}
    \rkhsvect_\onedata = \int \smallkernel(\latfunc_{\weights}(\onedata),.)\probfunc(\weights)d\weights
\end{align}
The RKHS norm of the embedding $\rkhsvect_\onedata$ is thus given by
\begin{align}
    ||\rkhsvect_\onedata||_{\rkhs_\smallkernel}^2 &= <\rkhsvect_\onedata,\rkhsvect_\onedata>_{\rkhs_\smallkernel} = \int \smallkernel(\latfunc_{\weights}(\onedata),\latfunc_{\weights'}(\onedata))\probfunc(\weights)\probfunc(\weights')d\weights d\weights'\\
    &= \mathbb{E}_{\weights,\weights' \sim \probfunc(\randweights)}[\smallkernel(\latfunc_{\weights}(\onedata),\latfunc_{\weights'}(\onedata))] = \smallkernel_{**} \label{eq:rkhsnorm_1}
\end{align}
This holds for any data point $\onedata$ and thus holds for any unlabeled data point $\onedata \in \alldata_U$.

Let $\rkhs_{\smallkernel}^{L} \simeq \rkhs_\smallkernel$ be the subspace $\rkhs_\smallkernel$ with basis vectors $\rkhsvect_{\onedata_{1}},\ldots,\rkhsvect_{\onedata_{\numlab}}$, the RKHS embeddings of the distributions in $\latentspace$ generated by the $\numlab$ labeled points $\onedata_1, \ldots, \onedata_{\numlab}$. The orthogonal projection matrix for this subspace is given by \citep{roy2014linear}
\begin{align}
    P_{\rkhs_{\smallkernel}^{L}} = \mathbf{L}(\mathbf{L}^T\mathbf{L})^{-1}\mathbf{L}^{T}
\end{align}
where the $i^{\text{th}}$ column of the matrix $\mathbf{L}$ is the vector $\rkhsvect_{\onedata_i}$. Therefore the RKHS norm of the orthogonal projection $\rkhsvect_{\onedata}^{L}$ of $\rkhsvect_\onedata$, $\onedata \in \alldata_U$ onto the subspace $\rkhs_{\smallkernel}^{L}$ is given by
\begin{align}
    ||\rkhsvect_{\onedata}^{L}||_{\rkhs_\smallkernel}^{2} = (\rkhsvect_{\onedata}^{L})^{T}P_{\rkhs_{\smallkernel}^{L}}^{T}P_{\rkhs_{\smallkernel}^{L}}\rkhsvect_{\onedata}^{L} = (\rkhsvect_{\onedata}^{L})^{T}\mathbf{L}(\mathbf{L}^T\mathbf{L})^{-1}\mathbf{L}^{T}\rkhsvect_{\onedata}^{L}
\end{align}
The $(i,j)^{\text{th}}$ entry of $\mathbf{L}^T\mathbf{L}$ is given by the inner product
\begin{align}
    <\rkhsvect_{\onedata_i},\rkhsvect_{\onedata_j}>_{\rkhs_\smallkernel} = \int \smallkernel(\latfunc_{\weights}(\onedata_i),\latfunc_{\weights'}(\onedata_j))\probfunc(\weights)\probfunc(\weights')d\weights d\weights'
\end{align}
which is precisely the $\kernel_{ij}$, the $(i,j)^{\text{th}}$ entry of the GP kernel matrix $\kernmat$. Similarly the $j^{\text{th}}$ entry of $\mathbf{L}^{T}\rkhsvect_{\onedata}^{L}$ is given by the inner product
\begin{align}
    <\rkhsvect_{\onedata},\rkhsvect_{\onedata_j}>_{\rkhs_\smallkernel} = \int \smallkernel(\latfunc_{\weights}(\onedata),\latfunc_{\weights'}(\onedata_j))\probfunc(\weights)\probfunc(\weights')d\weights d\weights'
\end{align}
which is the $j^{th}$ entry of the vector $\mathbf{\smallkernel_{*}}$ ($\mathbf{\smallkernel_{*}}[i] = \mathbb{E}_{\weights,\weights' \sim \probfunc(\randweights)}[\smallkernel(\latfunc_{\weights}(\onedata),\latfunc_{\weights'}(\onedata_i))]$). Thus overall
\begin{align}
    ||\rkhsvect_{\onedata}^{L}||_{\rkhs_\smallkernel}^{2} = \mathbf{\smallkernel_{*}}^{T}\kernmat^{-1}\mathbf{\smallkernel_{*}} \label{eq:rkhsnorm_2}
\end{align}
Finally from the Pythagoras Theorem we know that
\begin{align}
    ||\omega_{\onedata} - \omega_{\onedata}^{L}||_{\rkhs_\smallkernel}^2 = ||\omega_{\onedata} ||_{\rkhs_\smallkernel}^2 - ||\omega_{\onedata}^{L}||_{\rkhs_\smallkernel}^2 = \smallkernel_{**} - \mathbf{\smallkernel_{*}}^{T}\kernmat^{-1}\mathbf{\smallkernel_{*}} = \stdev^{2}(\onedata)
\end{align}
where the last equality is obtained by substituing the results of \Fref{eq:rkhsnorm_1} and \Fref{eq:rkhsnorm_2}. Note that in this case we assume that the RKHS embeddings $\rkhsvect_{\onedata_{1}},\ldots,\rkhsvect_{\onedata_{\numlab}}$ are linearly independent and form a basis for $\rkhs_{\smallkernel}^{L} \simeq \rkhs_\smallkernel$. 

\begin{figure*}[htb]
  \centering
   \subfloat[CTSlice ($\highdim = 384$)]{\includegraphics[width=0.33\linewidth]{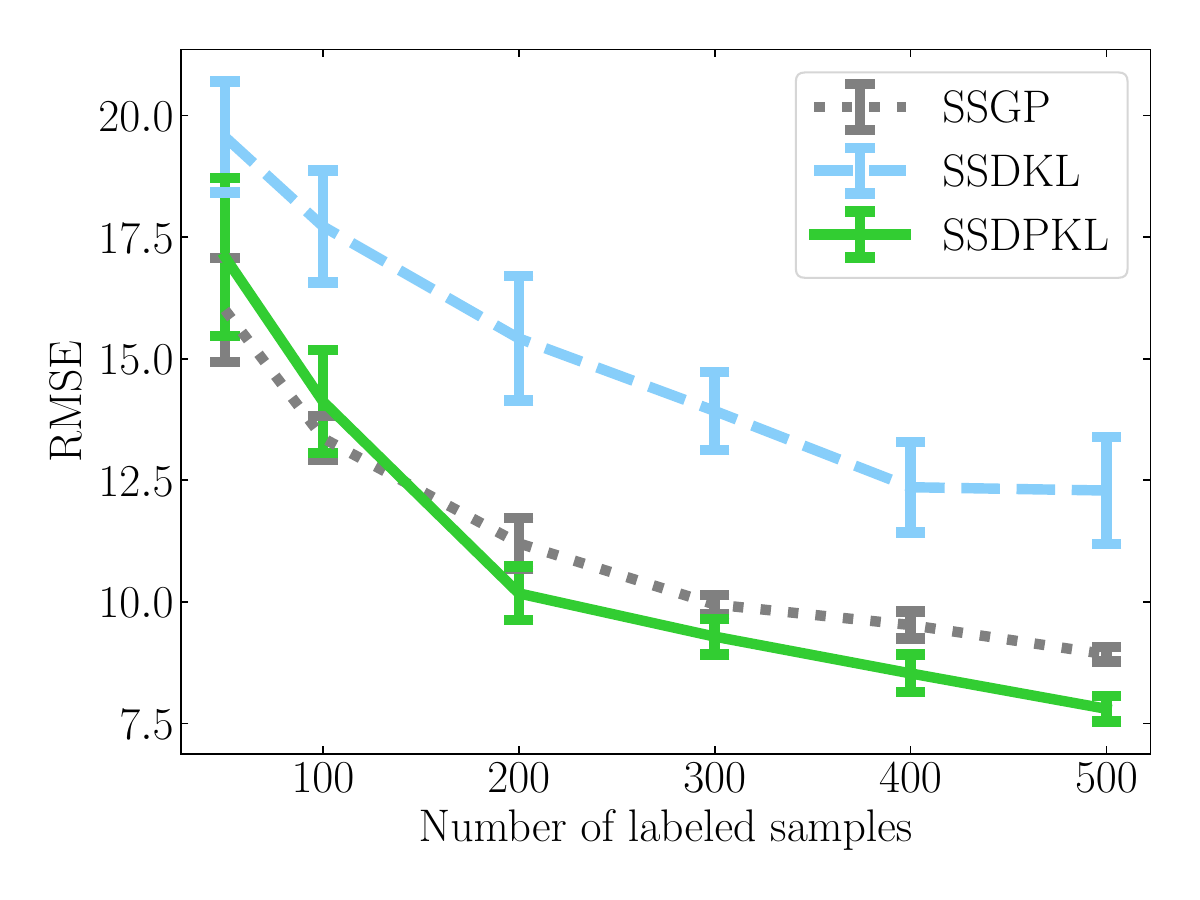}}
   \subfloat[Electric ($\highdim = 6$)]{\includegraphics[width=0.33\linewidth]{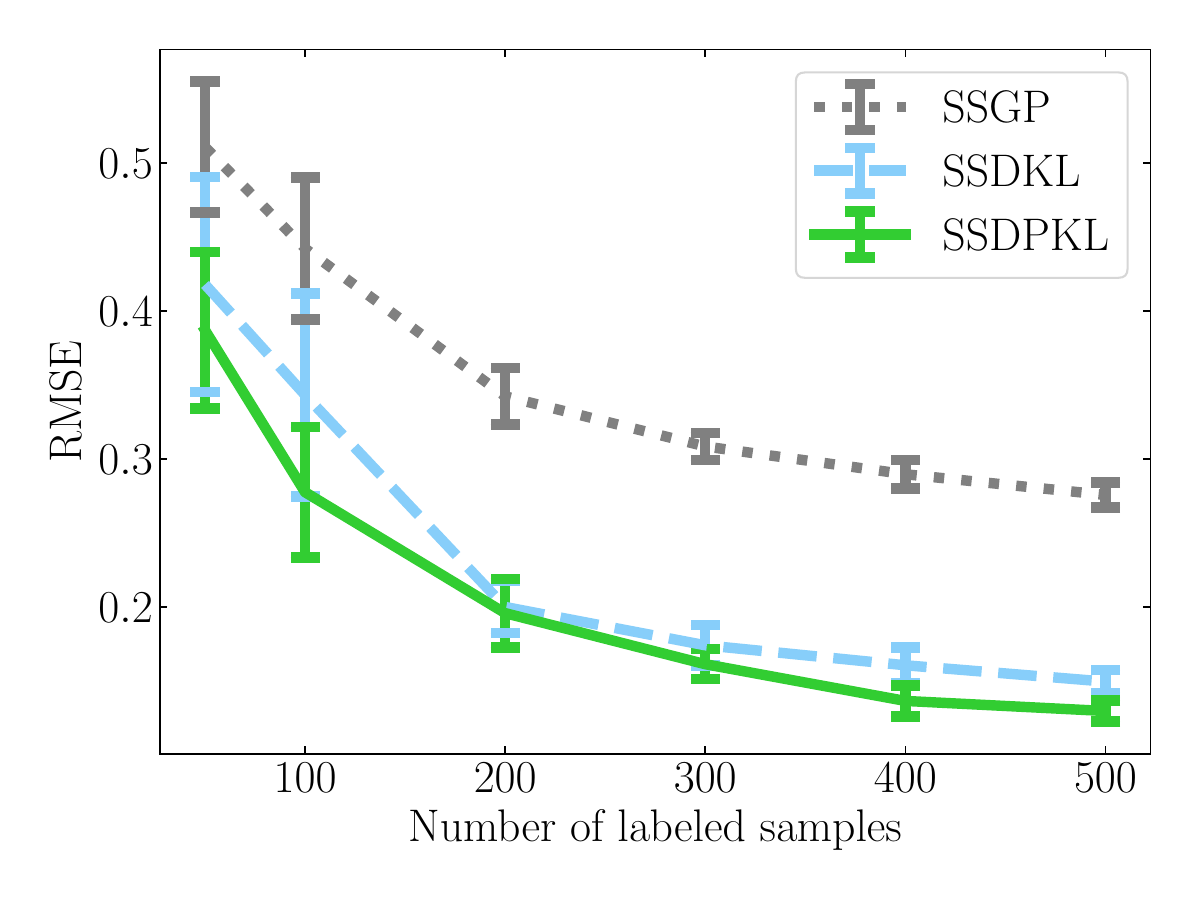}}
   \subfloat[Parkinsons ($\highdim = 20$)]{\includegraphics[width=0.33\linewidth]{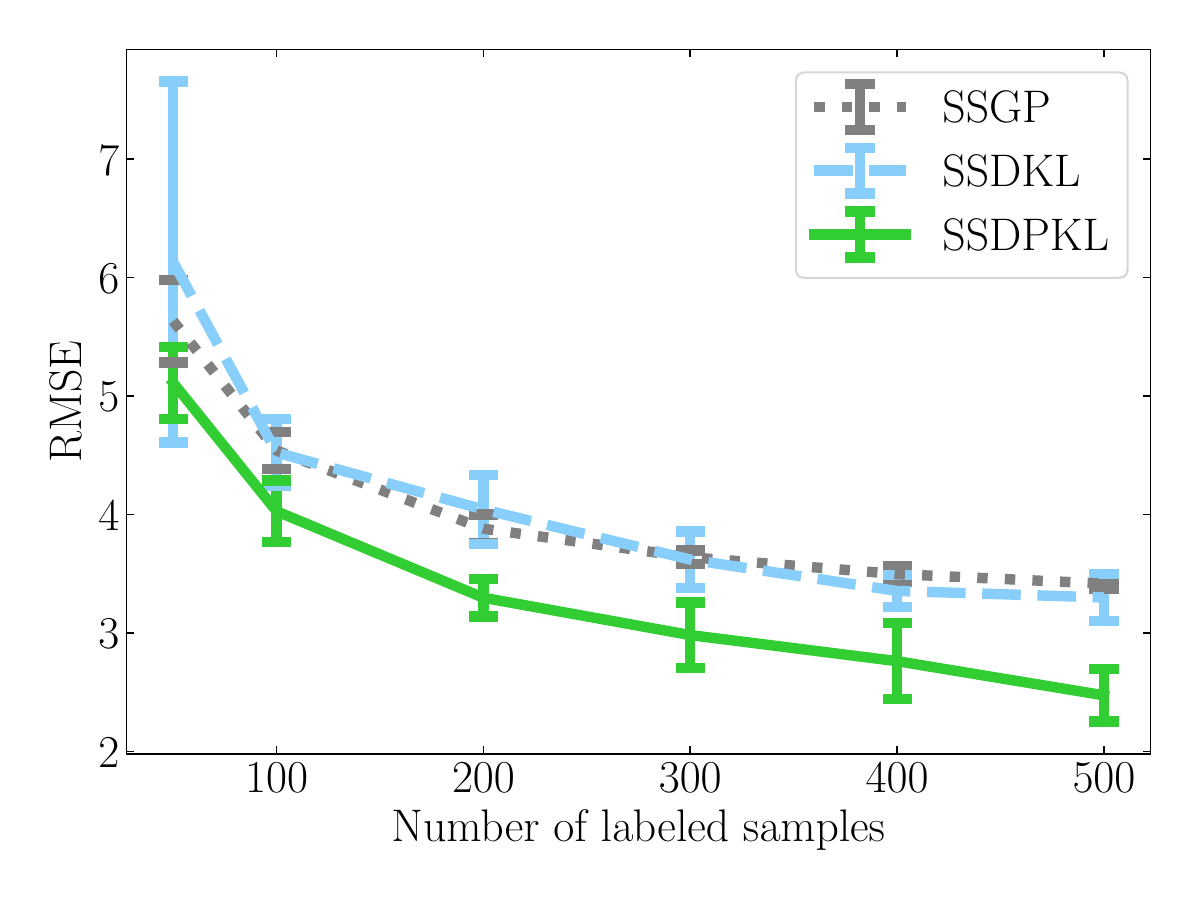}}\\
    \subfloat[Skillcraft ($\highdim = 18$)]{\includegraphics[width=0.33\linewidth]{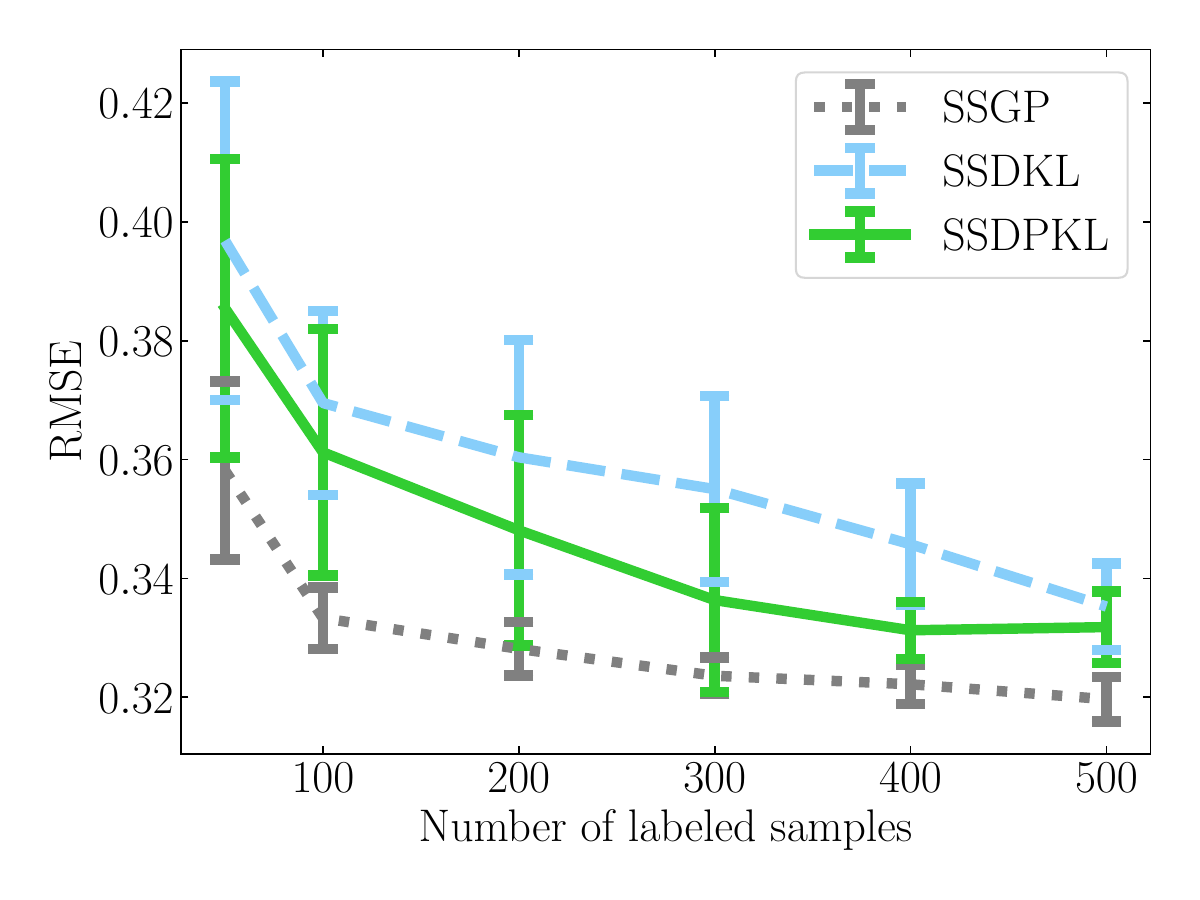}}
   \subfloat[Buzz ($\highdim = 77$)]{\includegraphics[width=0.33\linewidth]{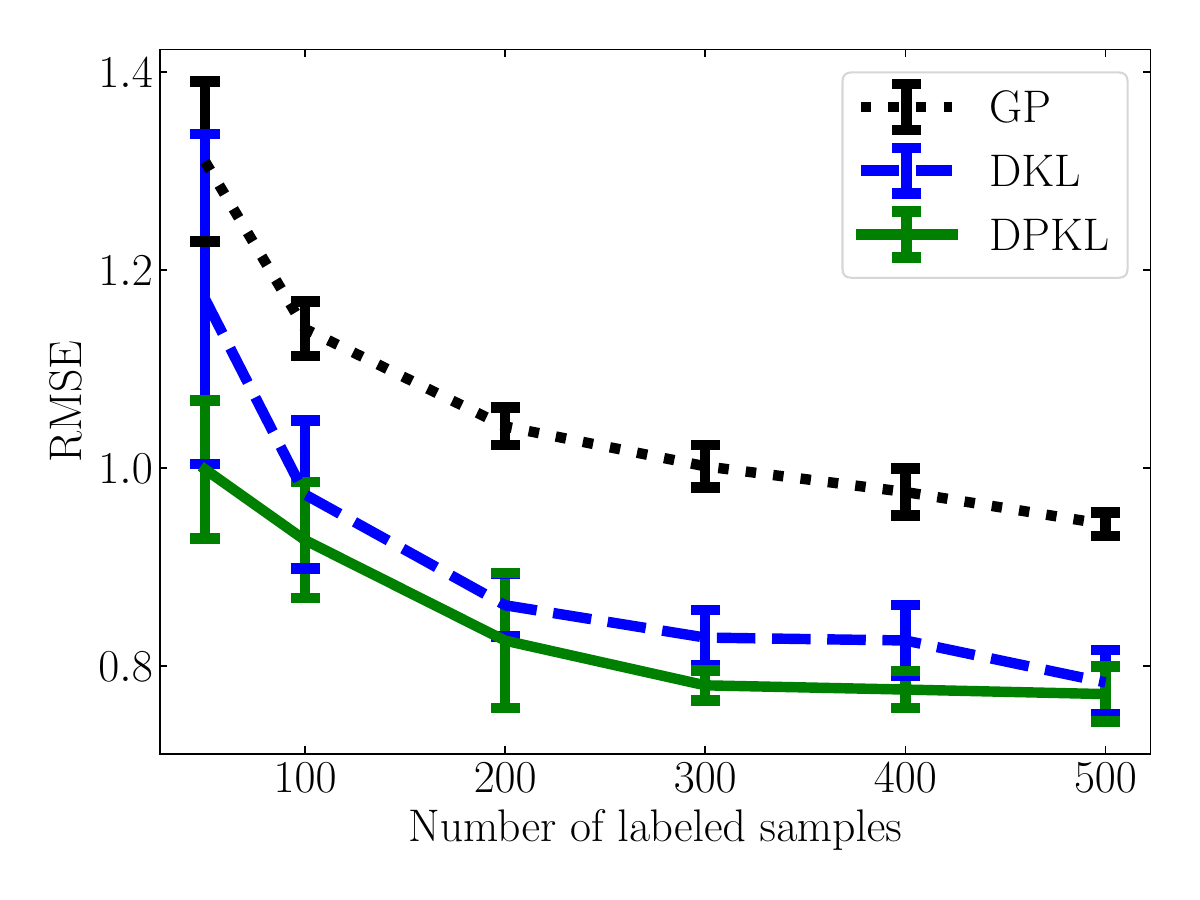}}
   \subfloat[Elevators ($\highdim = 18$)]{\includegraphics[width=0.33\linewidth]{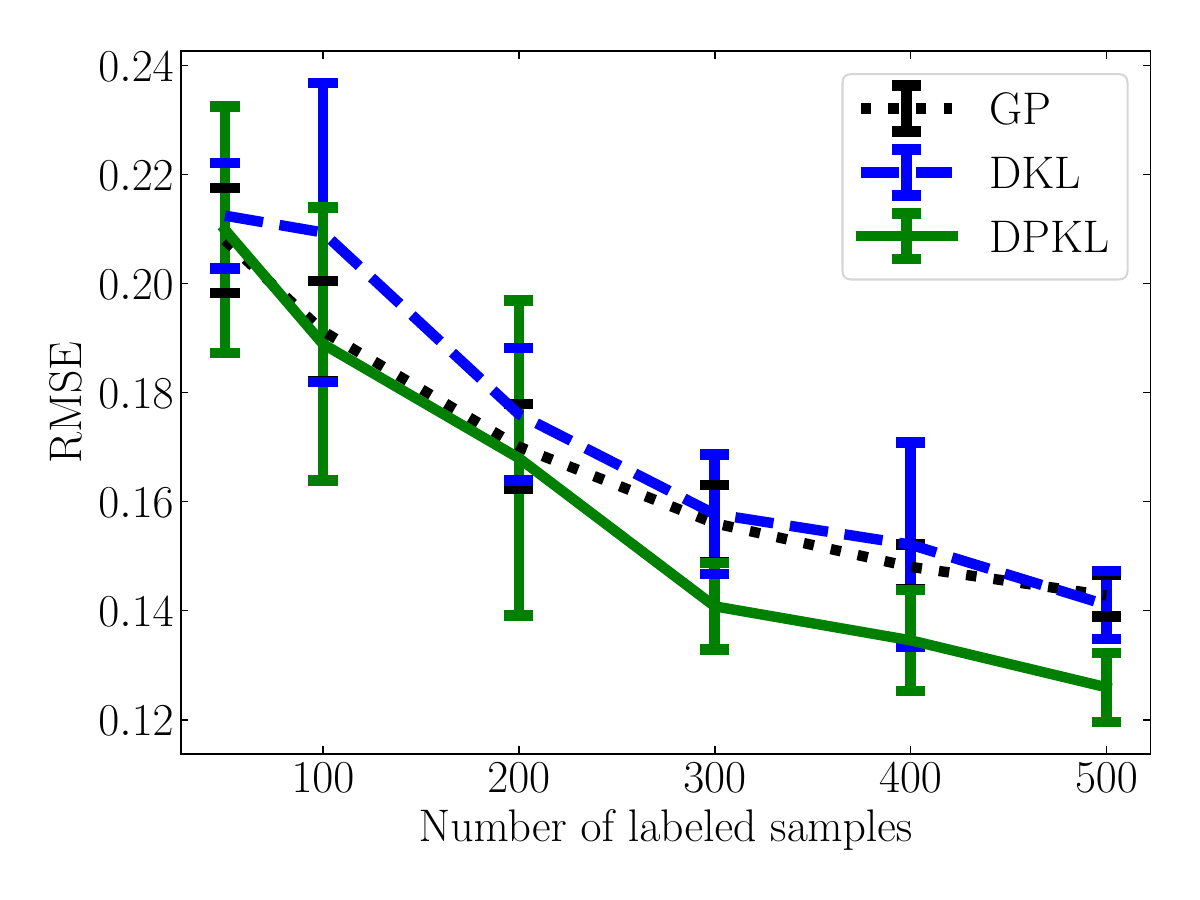}}
   \caption{Plots (a)-(d) show results for SSDPKL v/s SSDKL on 4 UCI datasets while Plots (e) and (f) show performance of DPKL v/s DKL on the other 2 datasets. DPKL and SSDPKL clearly outperform their deterministic counterpaprts - DKL and SSDKL in terms of test RMSE across all datasets.}
  \label{fig:UCI_app}
\end{figure*}
\section{Additional Experiments}
\label{sec:extra_app}
\subsection{UCI Regression} 
\Fref{fig:UCI_app} shows RMSE for semi-supervised models (SSDPKL/SSDKL/SSGP) on 4 datasets and for supervised models (DPKL/DKL/GP) on the other 2 datasets. These complement the plots in Figure 3 of the main paper. Once again we see that DPKL (SSDPKL) improves over DKL (SSDKL) across \emph{all} datasets. While SSGP does a bit better than SSDPKL on one of the lower dimensional datasets (Skillcraft), we note that SSDPKL beats SSGP on all other datasets, including the high dimensional datasets CTSlice ($D = 384$) and Buzz ($D = 77$), thus clearly illustrating the advantages of our approach in high dimensional settings.
\subsection{MNIST Classification} 
\begin{figure*}[htb]
  \centering
    \subfloat[Accurate Prediction]{\includegraphics[width=0.33\linewidth]{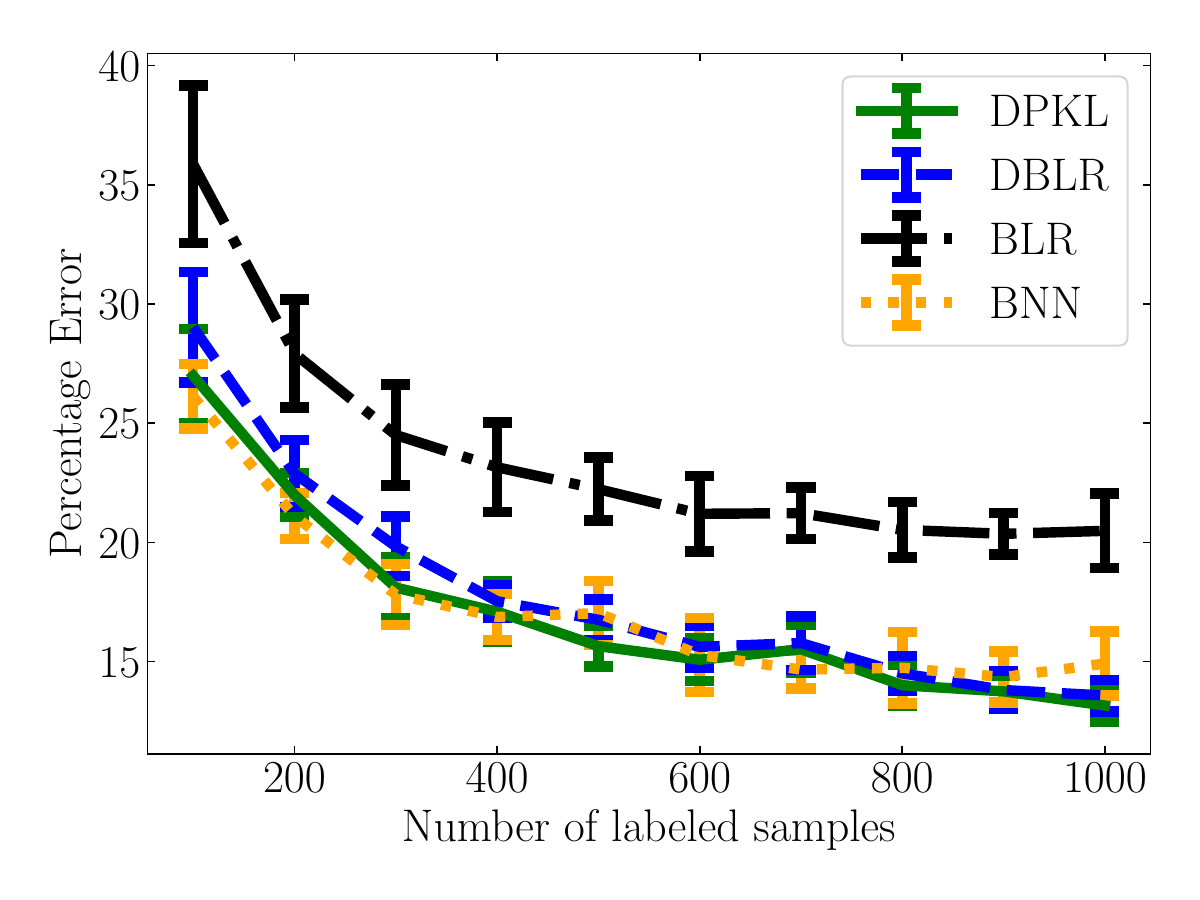}\label{fig:MNIST_acc}}
   \subfloat[Uncertainty Quantification]{\includegraphics[width=0.33\linewidth]{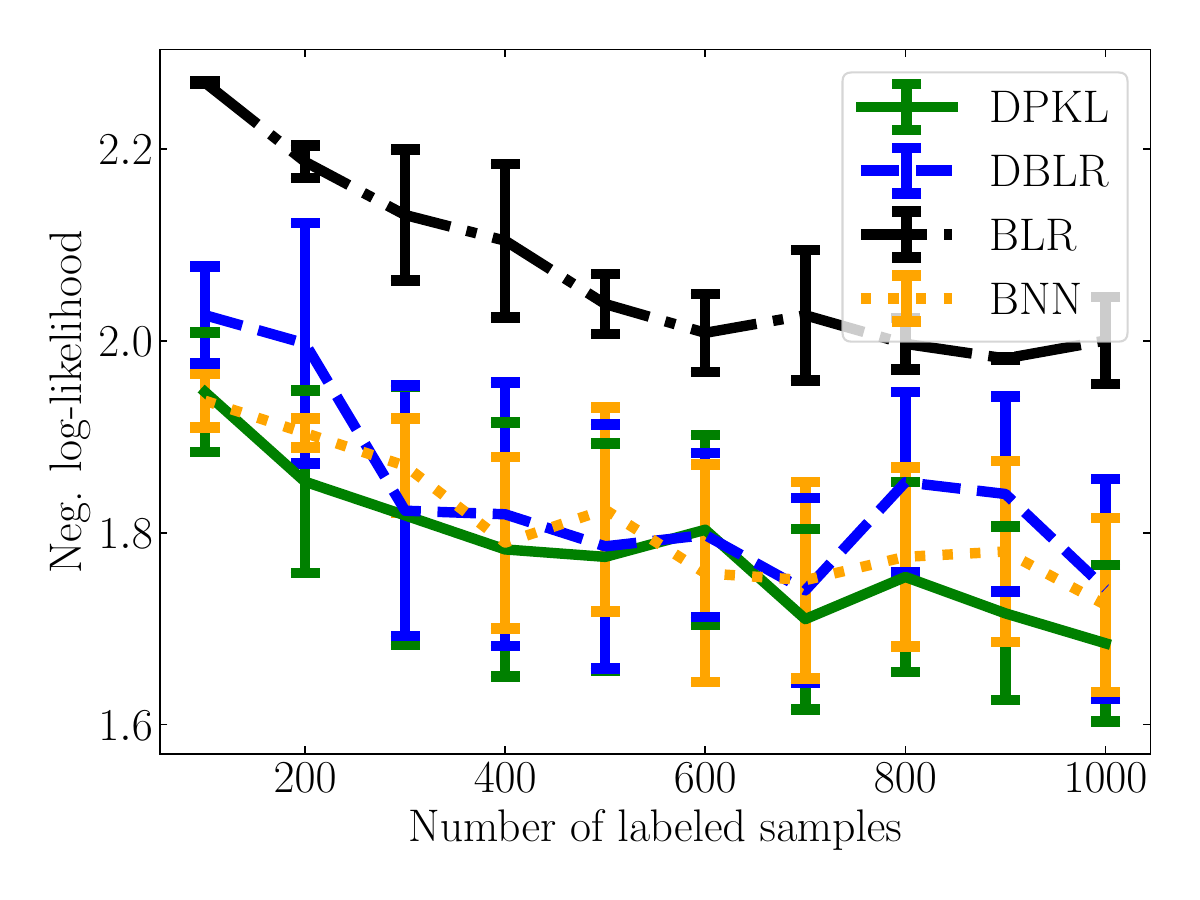}\label{fig:MNIST_uq}}
   \subfloat[Representation Learning]{\includegraphics[width=0.33\linewidth]{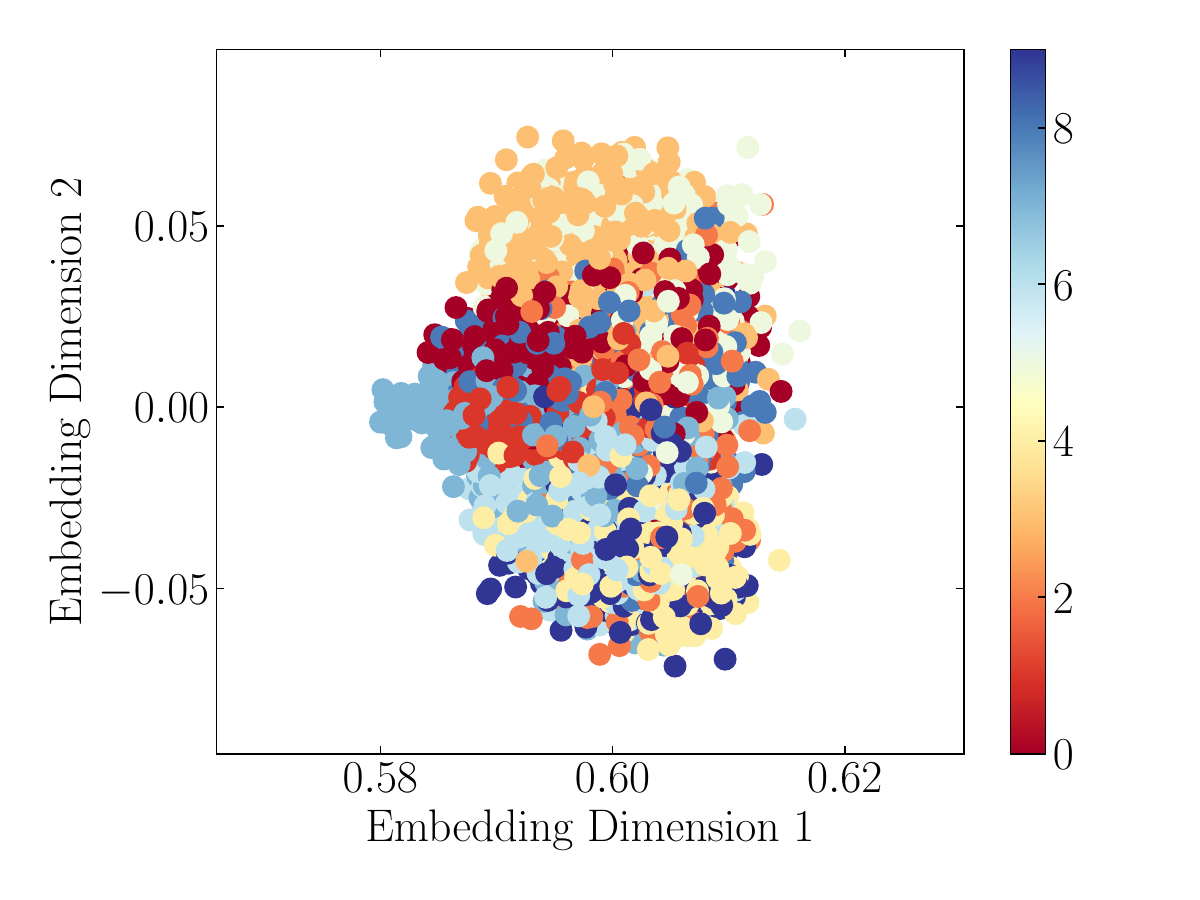}\label{fig:MNIST_rl}}
   \caption{(a) DPKL has lower classification error on MNIST as compared to baseline models with $100 \leq \numtrain \leq 1000$ labeled examples (b) The lower average negative log-likelihood of DPKL on test data indicates its superior uncertainty quantification over baseline models. (c) The average Random Fourier Feature embeddings of the learned latent distributions (with $\numtrain = 100$) show that DPKL learns a meaningful representation since points with the same label are close by in the latent space (the colorbar shows the digit label for each points) }
  \label{fig:MNIST}
\end{figure*}

We compare the extension of DPKL to classification (as described in Section 3.3 of the main paper) to 3 baselines, Bayesian Logistic Regression (BLR), Bayesian Logistic Regression on deterministic embeddings (similar to \citep{snoek2015scalable}) which we call DBLR, and a Bayesian Neural Network (BNN). Bayesian inference, where required, is performed using Stein Variational Gradient Descent \citep{liu2016stein}. The BNN learns the posterior over model parameters while DPKL learns the optimal distribution via maximum likelihood in the distribution space. All neural networks have a single hidden layer with $64$ nodes. We use $\numweights = 10$ for approximating expectations in all models. From the results in \Fref{fig:MNIST} it is clear that the deep models (DPKL, DBLR, and BNN) clearly improve over BLR both in terms of accuracy (lower classification error) and uncertainty quantification (lower negative log likelihood). The gains (lower error and negative log-likelihood) of DPKL over DBLR are not as pronounced as their regression counterparts (possibly due to low complexity of MNIST data). However as seen in the experiments on few-shot classification (Section 4) the gains are more significant on more complex data.

\section{Experiment Details}
\label{sec:details_app}
The Code for the experiments is available at \href{https://github.com/ankurmallick/DPKL}{https://github.com/ankurmallick/DPKL}. The experiments were run on machines with a 16-core Intel Xeon processor, 64 GiB RAM, TitanX GPU, 400GB NVMe SSD and a 2x4TB HDD. All experiments were run on CPU (GPU was not used).

\subsection{UCI Regression} 

\begin{table}[t]
\begin{center}
\begin{tabular}{ |c|c|c| } 
 \hline
 \textbf{Dataset} &  \textbf{Number of samples} & \textbf{Dimensionality}\\
 \hline
 \textbf{Buzz}  &  $583,250$ & $77$ \\ 
  \hline
 \textbf{CTSlice} & $53,500$ & $384$ \\ 
  \hline
 \textbf{Electric}  & $2,049,280$ & $6$ \\ 
  \hline
 \textbf{Elevators} & $16,599$ & $18$ \\ 
  \hline
 \textbf{Parkinsons} & $5,875$ & $20$ \\ 
  \hline
 \textbf{Skillcraft}  & $33,285$ & $18$ \\ 
  \hline
\end{tabular}
\end{center}
\caption{Details (total number of samples and dimensionality) of each dataset from the UCI repository, used for regression}
\label{tbl:UCI}
\vspace{-5mm}
\end{table} 

\textbf{Model Details: }  The GP and SSGP models use a Squared Exponential (SE) kernel directly on the data i.e. $\smallkernel(\onedata_i,\onedata_j) = \exp(-\frac{1}{2}\sum_{l}\frac{1}{h_{l}^{2}}(\onedata_{il} - \onedata_{jl})^2)$. The DKL and SSDKL models embed the data into a low-dimensional latent space before using an SE kernel i.e. $\smallkernel(\onedata_i,\onedata_j) = \exp(-\frac{1}{2}\norm{\latfunc_{\weights}(\onedata_i) - \latfunc_{\weights}(\onedata_j)})$. The DPKL and SSDPKL models embed the data into a \emph{probability distribution} in the latent space and then use the SE kernel between distributions (as described in Section 3 of the main paper). We used the same neural network architecture as \citep{jean2018semi} ($\highdim - 100 - 50 - 50 - \lowdim$) for mapping data points to the latent space in the DKL and DPKL models. Here $\highdim$ is the dimensionality of the datasets and $\lowdim = 2$ is the dimensionality of the latent embedding $\onelatent \in \latentspace$. In DPKL and SSDPKL, we used $\numweights = 10$ samples from this architecture to represent the distribution over model parameters. Following \citep{liu2016stein} we use the RBF kernel as the kernel $\kappa$ between model parameters $\weights$  (for functional gradient descent) with bandwidth chosen according to the median heuristic described in their work since it causes $\sum_{j}\wtkernel(\weights,\weights_j) \simeq 1$ for all $\weights$, leading $\wtkernel$ to behave like a probability distribution. The lengthscales for the GP SE kernel, and the neural network weights in DKL are optimized using gradient descent on the negative log-likelihood while the \emph{distribution} over neural network weights in DPKL is optimized using functional gradient descent (Algorithm 1). We used the Random Fourier Features Approximation as explained in Section 3.1 of the paper for DPKL and SSDPKL with $R=100$ samples from the Fourier Transform of $k$. The hyperparameter GP noise variance is set to $\stdev_{\eta}^2 = 1$ and the regularization parameter in SSDPKL and SSDKL, is set to $\alpha = 1$.

\textbf{Optimizer:} We used the Adam Optimizer \citep{kingma2014adam}, with a learning rate of $10^{-3}$ and Nesterov Momentum as implemented in Tensorflow \citep{abadi2016tensorflow} under the name Nadam, to estimate model parameters in DPKL, DKL, SSDPKL, and SSDKL, as well as the GP kernel lengthscales, by maximum likelihood. 

\textbf{Data Processing:} Following \citep{jean2018semi}, we evaluate our models on 7 datasets of the UCI repository. The details of the datasets are given in Table 1. For each dataset and each model, we set aside $1000$ examples as the test set. From the remaining examples we vary the number of labeled examples as $\numtrain = \{50, 100, 200, 300, 400, 500\}$. Labels (in $\alltarget$) are normalized to have zero mean and unit variance. The supervised models - DPKL, DKL, and GP, are trained \emph{only} on the labeled examples while for the Semi-Supervised models - SSDPKL, and SSDKL, we use (at most) $10000$ unlabeled examples from the rest of the data (excluding the test set.) All models are trained for $50$ epochs. For models, we create a validation set of $10\%$ of the $\numtrain$ training examples for early stopping (checked after every $10$ epochs). For every value of $\numtrain$, models are trained on a different set of $\numtrain$ examples in each trial (keeping the test set fixed). The process is repeated for $10$ trials and models are evaluated in terms of prediction accuracy and uncertainty quantification. 

\textbf{Prediction Accuracy:} We use RMSE as a measure of prediction accuracy. RMSE is computed between the labels of the test data set, and the predicted mean values of the test labels under the corresponding model. Low RMSE implies predicted labels are close to true labels. We plot average test RMSE v/s number of labeled examples for all datasets and all models. We also include error bars corresponding to one standard deviation. 

\textbf{Uncertainty quantification:} Following \citep{lakshminarayanan2017simple}, we use test data negative log-likelihood as a measure of uncertainty quantification. For a test data point $\onedata_{*}$ with normalized target value $\onetarget_{*}$, if the Gaussian Process predicted mean is $\mu(\onedata_{*})$ and predicted variance is $\stdev^2(\onedata_{*})$, then the negative log-likelihood is given by
\begin{align*}
    \text{Negll} = \frac{(\onetarget_{*} - \mu(\onedata_{*})^2}{2\stdev^2(\onedata_{*})} + \frac{1}{2}\log \stdev^2(\onedata_{*}) + \frac{1}{2} \log (2\pi)
\end{align*}
Low negative log-likelihood implies that the model explains the data distribution (i.e quantifies uncertainty) well. We plot average test negative log-likelihood for DPKL, DKL, and GP, for all datasets in \Fref{fig:UCI_app} with error bars corresponding to one standard deviation. 

\subsection{Few-shot Classification} 

In DPKL for few-shot classification, data is projected into distributions in a 128-dimensional latent space, by passing it through a neural network with a single hidden layer of $128$ nodes, followed by Bayesian Logistic Regression for prediction as described above. We evauate DPKL, DBLR, and the deterministic 1-NN classifier described in Appendix A4 of \citep{chen2019closer} on the CUB and mini-Imagenet datasets considered in their paper. As in \citep{chen2019closer}, the mini-Imagenet dataset consists of a subset of $100$ classes from the original Imagenet dataset. During training, a feature extractor (4-layer convolution backbone) is trained on all examples of $64$ classes called \emph{base classes} as per the Baseline++ approach in \citep{chen2019closer}. In the fine tuning stage $20$ of the remaining $36$ classes were selected as \emph{novel}. Each trial involves drawing a random set of $5$ novel classes. The predictive model (DPKL, DBLR, or 1-NN) is trained on the outputs of the feature extractor for $5$ examples per novel class ($25$ training points in total) and is evaluated on a query set consisting of $16$ examples per novel class. For the CUB dataset we have $100$ base classes and $50$ novel classes as per \citep{chen2019closer} with everything else remaining the same. For DPKL and DBLR we use $\numweights = 10$ particles to approximate distributions over model parameters. All models are trained for 100 epochs (minibatch size = 5), and we pick the parameters which give the least training error (checked after every 10 epochs). We report the accuracy on the query set, averaged over 600 trials, along with $95\%$ confidence intervals in Table 1 of the main paper.

\subsection{MNIST Classification}

\textbf{Model Details:} We use SVGD for performing Bayesian Inference in the Bayesian Logistic Regression (BLR) and Bayesian Neural Network (BNN) models as described in \citep{liu2016stein}, as well as in the output layer of the Deep BLR (DBLR), and DPKL models since this layer is performing Bayesian Logistic Regression. The parameters of the embedding (single layer neural network with $64$ nodes) are estimated via gradient descent and functional gradient descent (Algorithm 1) for DBLR and DPKL  respectively.  Since the latent dimensionality is higher in this case ($64$ as compared to $2$ for regression) we used Orthogonal Random Features \citep{yu2016orthogonal} with $R=640$ features, to approximate the kernel $\smallkernel$ in the latent space, since they are based on the same principle but give lower variance approximations than Random Fourier Features.

\textbf{Optimizer: } We used minibatches of size $32$ for calculating gradients in all models (as opposed to full batch gradient descent in regression) since the cross entropy loss in (17) admits minibatches. All models were trained using the Adam Optimizer \citep{kingma2014semi} with a learning rate of $10^{-3}$. The regularization parameter of SVGD was set to $10^{-3}$.

\textbf{Data Processing: } We compare classification models on $100 \leq \numtrain \leq 1000$ labeled examples from the MNIST training data in Tensorflow. We ensure that there are equal number of examples  of each class for all values of $\numtrain$.  Models are evaluated on the standard MNIST test data in Tensorflow. Each model was trained for 100 epochs (1 epoch corresponds to a pass over all $\numtrain$ data points) and the model with the best validation accuracy (by checking after every 10 epochs) was selected. We used a $90-10$ validation split for this purpose. The process is repeated for $10$ trials and models are evaluated in terms of prediction accuracy and uncertainty quantification.

\textbf{Prediction Accuracy:} Since the last (output) layer of each model performs Bayesian Logistic Regression, therefore the model outputs samples from the distribution over softmax probabilities for each test data point. Thus if $\numweights$ particles are used to estimate probabilities in the last layer then we have $\numweights$ samples of $\classprob_{i\classind}$, the probability that test point $i$ belongs to class $\classind$ for all $i,\classind$. The final prediction for point $i$ is the class $\classind$ with the highest average $\classprob_{i\classind}$, averaged over all $\numweights$ samples (we us $\numweights=10$ in our experiments). We report the average percentage classification error (averaged over 10 trials, error bars corresponding to one standard deviation) on the $10000$ MNIST test data points as a measure of prediction accuracy in Fig. 4a of the main paper.

\textbf{Uncertainty Quantification:} We use the test data negative log-likelihood for uncertainty quantification. The negative log likelihood for a data point $i$ is the cross entropy loss between the one-hot encoded vector corresponding to its class label, and the vector of average class probabilities $\classprob_{i\classind}$, averaged over all $\numweights$ samples of the output layer. We report the average test data negative log likelihood, as per (17), (averaged over 10 trials, error bars corresponding to one standard deviation) on the $10000$ MNIST test data points as a measure of prediction accuracy in Fig. 4b of the main paper.
\end{appendix}

\bibliography{Bibliography}
\end{document}